\documentclass[10pt,twocolumn,letterpaper]{article}

\usepackage[pagenumbers]{cvpr} % To force page numbers, e.g. for an arXiv version

\definecolor{cvprblue}{rgb}{0.21,0.49,0.74}
\usepackage[pagebackref,breaklinks,colorlinks,allcolors=cvprblue]{hyperref}
\usepackage{amsmath, amssymb}
\usepackage{afterpage}
\usepackage{xcolor}
\usepackage{colortbl}
\usepackage{multirow}
\usepackage{subcaption}
\usepackage{float} 

\usepackage{tcolorbox}
\tcbuselibrary{listings, breakable}

\newtcblisting{promptbox}[1][]{
    colback=gray!10,
    colframe=black,
    fonttitle=\bfseries\small,
    top=2pt, bottom=2pt, left=2pt, right=2pt,
    boxsep=5pt,
    arc=3pt, outer arc=3pt,
    listing only,
    listing options={
        gobble=30,
        keepspaces=false,
        columns=fixed,
        basewidth=0.5em,
        basicstyle=\ttfamily,
        breaklines=true,          % 启用自动换行
        breakindent=0pt,          % ✅ 关键：换行后不缩进，与首行对齐
        prebreak={},              % 可选：换行前不加符号
        postbreak={},             % 可选：换行后不加符号（如 ↪）
    },
    #1
}

\def\paperID{9119} % *** Enter the Paper ID here
\def\confName{CVPR}
\def\confYear{2026}

\newcommand{\tablestyle}[2]{\setlength{\tabcolsep}{#1}\renewcommand{\arraystretch}{#2}\centering\footnotesize}
\renewcommand{\paragraph}[1]{\vspace{1.25mm}\noindent\textbf{#1}}

\newcolumntype{x}[1]{>{\centering\arraybackslash}p{#1pt}}
\newcolumntype{y}[1]{>{\raggedright\arraybackslash}p{#1pt}}
\newcolumntype{z}[1]{>{\raggedleft\arraybackslash}p{#1pt}}

\newcommand{\app}{\raise.17ex\hbox{$\scriptstyle\sim$}}

\definecolor{deemph}{gray}{0.6}

\definecolor{baselinecolor}{gray}{.9}

\definecolor{dt}{HTML}{ADCAD8}
\definecolor{dt2}{HTML}{cddfe7}

\definecolor{defaultcolor}{HTML}{E8E2F7}

\let\cite\citep

\renewcommand{\paragraph}[1]{\vspace{1.25mm}\noindent\textbf{#1}}
\newlength\savewidth\newcommand\shline{\noalign{\global\savewidth\arrayrulewidth
  \global\arrayrulewidth 1pt}\hline\noalign{\global\arrayrulewidth\savewidth}}

\newcolumntype{x}[1]{>{\centering\arraybackslash}p{#1pt}}
\newcolumntype{y}[1]{>{\raggedright\arraybackslash}p{#1pt}}
\newcolumntype{z}[1]{>{\raggedleft\arraybackslash}p{#1pt}}
\definecolor{degray}{gray}{.6}

\title{Multi-Path Collaborative Reasoning via Reinforcement Learning}

\author{
Jindi Lv$^{1,2}$\quad
Yuhao Zhou$^{1}$\quad
Zheng Zhu$^{2}$\quad 
Xiaofeng Wang$^{2,3}$\quad
Guan Huang$^{2}$\quad
Jiancheng Lv$^{1}$\quad
\\
$^{1}$Sichuan University\;
\quad
$^{2}$GigaAI\;
\quad
$^{3}$Tsinghua University\;\\
Project page: \href{https://multi-path-collaborative-reasoning.github.io/}{https://multi-path-collaborative-reasoning.github.io/}
}

\begin{document}

\twocolumn[{
\renewcommand\twocolumn[1][]{#1}
\maketitle
\begin{center}
\centering
    \captionsetup{type=figure, margin=2pt}
    % 设置子标题编号的位置是120pt
    \subfloat[Average scores across five knowledge datasets: GRPO vs.\ M3PO.
]{
        \label{subfig:knowledge_m3po}
        \centering
        \includegraphics[width=0.225\linewidth]{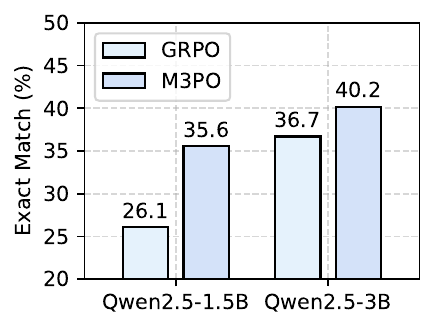}
    }
    \hspace{1pt}
    \subfloat[Average accuracy across five STEM datasets: GRPO vs.\ M3PO.
]{
        \label{subfig:stem_m3po}
        \centering
        \includegraphics[width=0.225\linewidth]{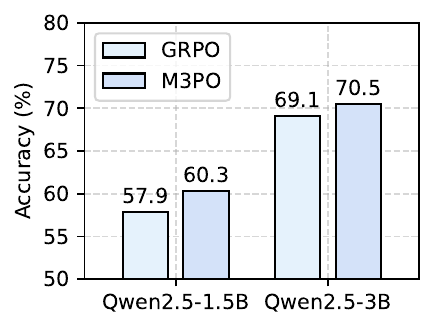}
    }
    \hspace{1pt}
    \subfloat[Training-free accuracy on MATH-500: Original vs.\ Soft Thinking.
]{
        \label{subfig:math_st} 
        \centering     
        \includegraphics[width=0.225\linewidth]{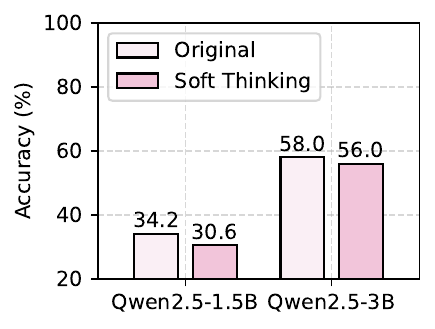}
    }
    \hspace{1pt}
    \subfloat[Training-free accuracy on GSM8k: Original vs.\ Soft Thinking.]{
        \label{subfig:gsm8k_st} 
        \centering     
        \includegraphics[width=0.225\linewidth]{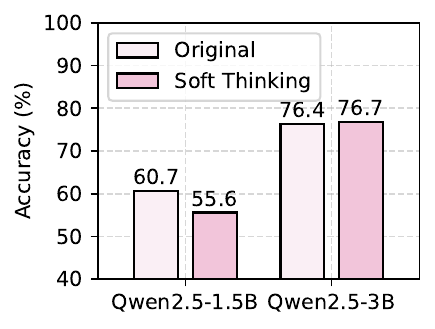}
    }
    \caption{Comparative analysis of reasoning paradigms. Figures (a) and (b) present results on knowledge and STEM benchmarks, while Figures (c) and (d) evaluate Soft Thinking without training on MATH-500 and GSM8k. Soft Thinking shows degraded performance, indicating extraneous inference noise. In contrast, M3PO consistently improves performance on knowledge- and reasoning-intensive benchmarks (with up to a \textbf{9.5\%} average gain on five knowledge datasets), demonstrating the superiority of multi-path reasoning paradigm.
    }
    
    \label{fig:top_figure}
\end{center}
}]

\begin{abstract}
Chain-of-Thought (CoT) reasoning has significantly advanced the problem-solving capabilities of Large Language Models (LLMs), yet conventional CoT often exhibits internal determinism during decoding, limiting exploration of plausible alternatives. Recent methods attempt to address this by generating soft abstract tokens to enable reasoning in a continuous semantic space. However, we find that such approaches remain constrained by the greedy nature of autoregressive decoding, which fundamentally isolates the model from alternative reasoning possibilities. In this work, we propose Multi-Path Perception Policy Optimization (M3PO), a novel reinforcement learning framework that explicitly injects collective insights into the reasoning process. M3PO leverages parallel policy rollouts as naturally diverse reasoning sources and integrates cross-path interactions into policy updates through a lightweight collaborative mechanism. This design allows each trajectory to refine its reasoning with peer feedback, thereby cultivating more reliable multi-step reasoning patterns. Empirical results show that M3PO achieves state-of-the-art performance on both knowledge- and reasoning-intensive benchmarks. Models trained with M3PO maintain interpretability and inference efficiency, underscoring the promise of multi-path collaborative learning for robust reasoning.

\end{abstract}    
\section{Introduction}
\label{sec:intro}
The advent of Chain-of-Thought (CoT)~\cite{wei2022chain,jaech2024openai,guo2025deepseek} has markedly improved the reasoning capabilities of Large Language Models (LLMs)~\cite{openai2023gpt,wei2022emergent,touvron2023llama} on complex tasks. By generating explicit intermediate steps in natural language, CoT decomposes problems and guides models step-by-step toward the answer~\cite{cho2022advances}. 
However, the discrete token decoding underlying standard CoT induces internal determinism, which limits the exploration of plausible alternatives~\cite{yao2023tree}.

Recent efforts seek to transcend discrete token decoding by introducing continuity into the reasoning process. Latent reasoning~\cite{hao2024training,shen2025codi,tack2025llm} propagates hidden states instead of tokens to enable differentiable reasoning, but such representations often deviate from the pretrained embedding space, requiring costly alignment and limiting compatibility~\cite{shen2025efficient}. To address this, Soft Thinking~\cite{zhang2025soft,wu2025llms} performs soft aggregation within the input embedding space, preserving model compatibility while simulating a smooth semantic flow.

\begin{figure}[t]
    \centering
    \includegraphics[width=1\linewidth]{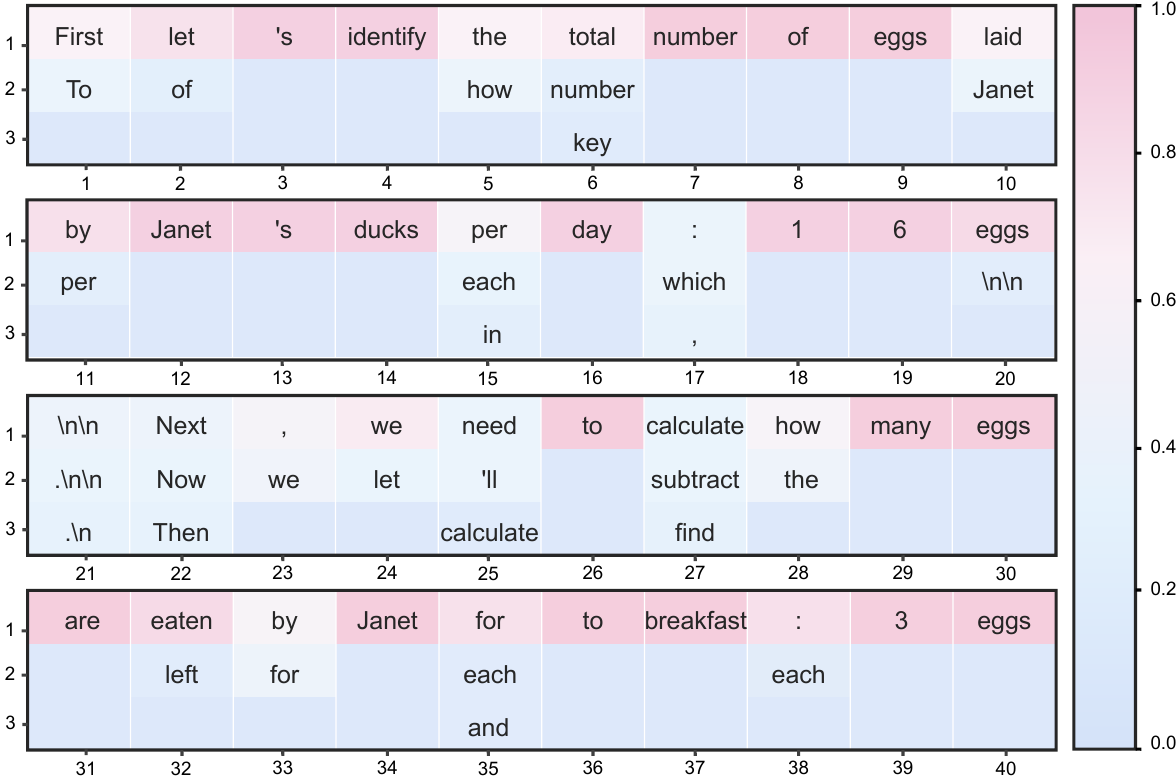}
    \caption{An illustrative case of probability distribution in vanilla Soft Thinking method. Each step aligns closely with the top-1 token from the prior step, leading to rapid concentration along a single semantic path.}
    \label{fig:st_case}
    \vspace{-10pt}
\end{figure}

Nevertheless, we reveal that Soft Thinking inherently lacks the ability to represent diverse semantic trajectories in parallel~\cite{wu2025llms}. As shown in Figure ~\ref{fig:st_case}, during consecutive decoding, each subsequent distribution consistently follows the dominant semantic token from the previous step, resulting in a coherent but exclusive progression along a single trajectory. 
Although soft aggregation improves information capacity, it primarily reinforces the dominant path while introducing extraneous noise over time, as evidenced in Figures \ref{subfig:math_st} and \ref{subfig:gsm8k_st}.

This phenomenon stems from the greedy nature of LLMs in autoregressive generation, where hidden states evolve at each step toward the most confident semantic direction~\cite{wu2025llms}. Such locally optimal decisions are amplified over long reasoning chains, resulting in a cumulative bias toward dominant paths and exclusion of alternatives. Thus, when a reasoning trajectory begins with a flawed premise, the lack of timely corrective feedback allows erroneous logic to propagate unimpeded through subsequent steps~\cite{xiong2025enhancing}.

In this work, we propose \textbf{M3PO}, a novel \textbf{M}ulti-\textbf{P}ath \textbf{P}erception \textbf{P}olicy \textbf{O}ptimization framework for reasoning robustness. 
M3PO leverages the inherent independence among policy rollouts to create natural multi-path sources.
Building on this perspective, we design a collaborative mechanism that explicitly incorporates cross-path interactions into the policy update. Through reward-guided policy optimization, M3PO progressively internalizes more consistent and reliable reasoning patterns.

This paradigm promotes robust policy optimization by allowing trajectories to benefit from cross-path insights. Specifically, the parameter-free gate facilitates step-level information exchange, providing trajectories with opportunities to gently refine their reasoning chain and reduce the persistence of local biases. In particular, by maintaining and refining multiple hypotheses in parallel, M3PO aligns with human-like cognition~\cite{fedorenko2024language,fedorenko2016language,benn2023language}, striking an adaptive balance between exploration and exploitation.

Empirically, as shown in Figures \ref{subfig:knowledge_m3po} and \ref{subfig:stem_m3po}, M3PO achieves superior performance over the standard CoT approach across both knowledge- and reasoning-intensive benchmarks. The advantage is particularly pronounced on knowledge-oriented tasks, where it delivers an average improvement of 9.5\%. These results validate the distinct advantage of our proposed multi-path reasoning paradigm.

We highlight the main contributions of this paper below:
\begin{itemize}
\item We hypothesize parallel rollouts in reinforcement learning as a natural source of reasoning diversity, eliminating the need for curated auxiliary datasets.
\item We design a multi-path collaborative learning mechanism that integrates cross-path interactions into policy updates, enabling implicit exploration-exploitation balance.
\item Our framework achieves state-of-the-art reasoning performance (up to +9.5\% accuracy over conventional CoT) on standard benchmarks without additional parameters.
\end{itemize}

\section{Related Works}
\label{sec:realted_works}
\subsection{Continuous Space Reasoning}
Recent works have explored continuous space reasoning as an alternative to explicit CoT generation~\cite{pfau2024let,goyal2023think,wang2023guiding,wang2023towards,turpin2023language}. Several studies demonstrate that Transformers can perform multi-hop reasoning implicitly through their internal hidden states, without generating intermediate textual steps~\cite{yang2024large,shalev2024distributional}. Deng et al.~\cite{deng2023implicit,deng2024explicit} further exploit this idea by using hidden dynamics directly for reasoning, bypassing symbolic output altogether. Geiping et al.~\cite{geiping2025scaling} extend this paradigm with a depth-recurrent architecture that reuses Transformer layers during inference, increasing computational depth per token and enabling internal processing.

Another line focuses on explicit continuous reasoning through latent representations~\cite{deng2023implicit,tack2025llm}. COCONUT~\cite{hao2024training} enables reasoning in the hidden state space by replacing discrete tokens with continuous vectors, eliminating the need for human-readable chains. CODI~\cite{shen2025codi} learns to align recurrent state updates via self-distillation, forming structured reasoning trajectories in the latent space. Although conceptual advances, they often struggle to scale to larger models and maintain performance on complex reasoning tasks.

\begin{figure*}[t!]
    % \centering
    % % \includegraphics[width=1\linewidth]{figures/crop_M3PO_pipeline_final_v5.pdf}
    \includegraphics[width=1\linewidth]{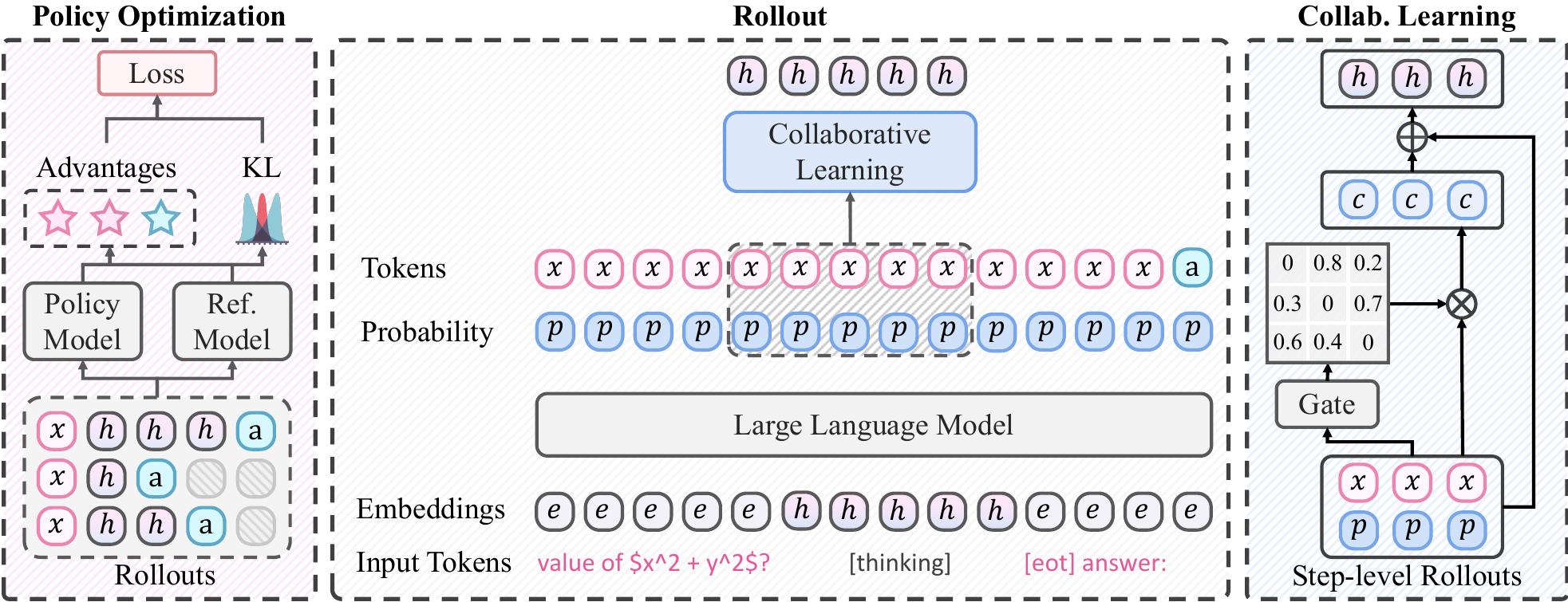}
    \caption{Overview of the M3PO framework. For a given question, multiple trajectories are first generated in parallel (Rollout). During thinking, these paths dynamically interact via a novel collaborative mechanism to refine intermediate reasoning steps (Collaborative Learning). The optimized trajectories are then used for policy updates via group-relative advantage estimation, closing the loop between multi-path exploration and policy learning (Policy Optimization).
}
    \label{fig:M3PO_pipeline}
    \vspace{-10pt}
\end{figure*}

More recently, training-free approaches such as Soft Thinking~\cite{zhang2025soft,gozeten2025continuous,wu2025llms} and Mixture-of-Inputs (MoI)~\cite{zhuang2025text} have emerged, leveraging output distributions to align hidden states with input embeddings, facilitating iterative reasoning in continuous space. Building on this, HRPO~\cite{yue2025hybrid} introduces a hybrid policy that fuses discrete token embeddings with soft embeddings through a gating mechanism, enabling richer internal state evolution during reasoning.

Despite these innovations, Soft Thinking methods mainly increase representational capacity without enabling meaningful exploration across reasoning paths. We propose a multi-path collaborative policy optimization framework that leverages parallel rollouts during training to encourage interaction among trajectories. By learning from alternative reasoning paths, the model breaks self-reinforcing logical loops and develops more robust inference capabilities.

\subsection{Reinforcement Learning}
Reinforcement learning (RL) has emerged as a powerful framework for enhancing language models through feedback-driven decision making. In natural language generation, RL enables models to learn from implicit human preferences by maximizing cumulative rewards~\cite{ouyang2022training}. This is typically achieved with policy gradient methods such as REINFORCE~\cite{sutton1999policy}, which update the policy using sampled responses and their rewards. Actor-critic approaches like A2C~\cite{mnih2016asynchronous} reduce variance with learned baselines, while PPO~\cite{schulman2017proximal} improves stability through a clipped objective that limits policy changes and prevents performance collapse.

An alternative line of work uses preference-based optimization, where models are fine-tuned on pairwise response comparisons~\cite{shakya2023reinforcement,chaudhari2025rlhf,zhang2025survey,wang2022deep}. Methods like DPO~\cite{rafailov2023direct} bypass explicit reward modeling by directly optimizing for preferred outputs, often eliminating the need for a reference model. Yet despite their efficiency, offline approaches generally lag behind online RL in complex reasoning settings. Recent online methods such as RLOO~\cite{ahmadian2024back}, GRPO~\cite{shao2024deepseekmath}, and REINFORCE++~\cite{hu2025reinforce++} address variance and scalability by computing baselines from multiple responses in the same batch. Leveraging group-level comparisons, they achieve more stable and memory-efficient training, making them well suited for long-horizon reasoning tasks.

Building on these advances, we propose an online RL framework that incorporates cross-path interactions during policy updates. By evaluating reasoning trajectories against each other within groups, the model learns to identify and correct flawed reasoning. This strengthens policy robustness by internalizing reliable reasoning patterns, yielding more accurate and stable text generation.

\section{Method}
\label{method}

\subsection{Overall Framework}
The M3PO framework establishes a closed-loop optimization process that seamlessly integrates multi-path reasoning with a step-level collaborative mechanism. As illustrated in Figure \ref{fig:M3PO_pipeline}, the framework comprises three core components: a parallel rollout pipeline, a collaboration learning mechanism, and a policy optimization module.

Formally, let $V$ be the vocabulary of size $|V|$ and $d$ be the embedding dimension. Given an input question $x$ and a token embedding matrix $E \in \mathbb{R}^{|V| \times d}$, the policy $\pi_\theta$ generates $N$ parallel rollouts. Each trajectory $\tau_i$ is represented as a sequence of embeddings:
\begin{equation}
    \tau_i = [E(x), \bar{h}_i^{(1)},  \dots, \bar{h}_i^{(L)}, E(a_i)], \quad i = 1, 2, \dots, N
\end{equation}
where $a_i$ denotes the final answer token, and $\bar{h}_i^{(l)}$ represents the hybrid thinking embedding at step $l$, produced by the collaborative mechanism.

This mechanism is activated only during the thinking phase. In contrast, the answer phase remains isolated to preserve the independence of final decisions. At each step $l$, the model outputs token distributions $p_i^{(l)} = \pi_\theta(\cdot \mid \tau_i^{(l)})$ for $i = 1, \dots, N$. The set $\{p_1^{(l)}, \dots, p_N^{(l)}\}$ is fed into a parameter-free gating function $\mathcal{G}$, which computes cooperation weights based on distributional consistency and value estimates. These weights are used to aggregate cross-path information and produce the hybrid thinking embedding $\bar{h}_i^{(l)}$, effectively fusing global insights into local trajectory.

Finally, the policy optimization module uses the collaboratively refined rollouts $\{\tau_1, \dots, \tau_N\}$ to compute advantages and update $\pi_\theta$ using a group-relative objective. This closed-loop design leverages reward signals to guide cross-path collaboration, ensuring that information exchange amplifies high-reward reasoning patterns while suppressing erroneous trajectories, thereby improving thinking quality and learning stability.

\subsection{Multi-Path Collaborative Mechanism}
The multi-path collaborative mechanism addresses the challenge of self-reinforcing reasoning loops by strategically introducing external perspectives during the thinking phase. This framework facilitates controlled information exchange across parallel trajectories, enabling each path to potentially break away from its own reasoning constraints while maintaining training stability.

At each thinking step \(l\), for the \(i\)-th trajectory with its sampled token embedding \(e_i^{(l)}\) and  the cross-path contextual embedding  \(c_i^{(l)}\), the mechanism constructs a hybrid thinking embedding through convex combination:
\begin{equation}
\label{eq:hybrid}
    \bar{h}_i^{(l)} = (1-\lambda)\,e_i^{(l)} + \lambda \,c_i^{(l)}, \quad \lambda \in [0,1]
\end{equation}
The hyperparameter $\lambda$ balances two crucial aspects, with the $(1-\lambda)$ term preserving the trajectory's intrinsic reasoning direction and the $\lambda$ term incorporating diversified information from alternative paths. This design allows each trajectory to maintain its distinctive characteristics while gaining opportunities to refer other reasoning patterns.

The cross-path contextual embedding \(c_i^{(l)}\) is generated through a distribution-similarity fusion mechanism. The process begins with computing a similarity matrix from output distributions:
\begin{equation}
S_{ij}^{(l)} = \frac{p_i^{(l)} \cdot p_j^{(l)}}{\|p_i^{(l)}\| \|p_j^{(l)}\|}, \quad i,j=1,\dots,N
\end{equation}
Diagonal elements are masked to prevent self-reinforcement:
\begin{equation}
\tilde{S}_{ij}^{(l)} =
\begin{cases}
0 & i=j, \\
S_{ij}^{(l)} & \text{otherwise}.
\end{cases}
\end{equation}
Cooperative weights are obtained through temperature-scaled normalization:
\begin{equation}
\label{eq:temp_softmax}
A_{ij}^{(l)} =
\begin{cases}
0 & i=j, \\
\displaystyle \frac{\exp(\tilde{S}_{ij}^{(l)}/T)}{\sum_{k\neq i}\exp(\tilde{S}_{ik}^{(l)}/T)} & \text{otherwise},
\end{cases}
\end{equation}
yielding the final contextual embedding:
\begin{equation}
c_i^{(l)} = \sum_{j=1}^N A_{ij}^{(l)} e_j^{(l)} \in R^d.
\end{equation}
This fusion mechanism ensures moderated information exchange by fostering strong interactions between distributionally aligned paths to smooth the evolution of reasoning states, while permitting controlled diversity through weaker engagements with divergent trajectories to prevent disruptive perturbations. 
% This collaborative process provides refined reasoning trajectories for the subsequent global policy optimization, enabling the identification and reinforcement of more robust reasoning strategies.

The resulting hybrid embedding $\bar{h}_i^{(l)}$ is used as the input for the next reasoning step, thereby directly shaping the model's subsequent reasoning and token predictions through integrated individual reasoning and collective insights. Notably, the mechanism operates exclusively during thinking steps, while the final answer generation remains independent to preserve decision diversity. When a trajectory exits the thinking mode at some step, it no longer participates in subsequent cross-path collaboration. Before the softmax normalization, we set $\tilde{S}_{ir}^{(l)}=\tilde{S}_{ri}^{(l)}=0$ for all $i$ for the exited trajectory $r$, so that it neither contributes to nor receives information in later steps.
%这里还要强调一下当一个trajectory停止thinking mode的时候，他要被掩码掉，不做交互

Through this local collaborative process, the mechanism provides refined reasoning trajectories for the subsequent global policy optimization, enabling the identification and reinforcement of more robust reasoning strategies.

\subsection{Policy Optimization}
M3PO optimizes the policy model $\pi_\theta$ via reinforcement learning, leveraging the intrinsic reasoning capabilities of LLMs without explicit supervision and curated auxiliary datasets. Building upon the refined thinking embeddings produced by a collaborative mechanism, we formulate the optimization objective to leverage both explicit reward signals and implicit guidance from cross-path interactions.

Formally, given an input question $x$ and a set of $N$ refined rollouts $\{\boldsymbol{\tau}_1, \boldsymbol{\tau}_2, \ldots, \boldsymbol{\tau}_N\}$ generated through our collaborative mechanism, we compute trajectory-level rewards $R(\boldsymbol{\tau}_i)$ based on final answer correctness. The advantage for each trajectory is estimated through group-relative standardization:
\begin{equation}
    A(\boldsymbol{\tau}_i) = \frac{R(\boldsymbol{\tau}_i) - \mu}{\sigma}
\end{equation}
where $\mu$ and $\sigma$ represent the mean and standard deviation of rewards within the group of $N$ rollouts. This advantage estimation provides a stable learning signal without requiring a separate value function.

The policy optimization objective combines advantage-weighted likelihood maximization with KL-divergence regularization:
\begin{equation}
\label{eq:objective}
\begin{split}
\nabla_\theta J_{\text{M3PO}}(\theta) &= 
\mathbb{E} \Bigg[ \frac{1}{N} \sum_{i=1}^N \bigg( \sum_{t=1}^{L} \nabla_\theta \log \pi_\theta(e_i^{(t)} \mid x, \bar{h}_i^{(<t)}) \\
&\quad \cdot A(\boldsymbol{\tau}_i) \bigg) \Bigg] - \beta \nabla_\theta D_{\text{KL}}[\pi_\theta \|\pi_{\text{ref}}],
\end{split}
\end{equation}
where $\beta$ controls the strength of the KL constraint, and $\pi_{\text{ref}}$ denotes the reference model for regularization. Crucially, the gradient updates are computed using the hybrid thinking embeddings $\bar{h}_i^{(<t)}$ as context, ensuring that the policy learns to leverage the collaborative reasoning patterns cultivated during the thinking phase. 

Note that M3PO uses raw log probabilities in its policy gradient (Equation~(\ref{eq:objective})) rather than likelihood ratios as in PPO~\cite{schulman2017proximal} or GRPO~\cite{shao2024deepseekmath}, eliminating the need for ratio clipping under our conservative update schedule. Furthermore, each trajectory is used only once for gradient updates since the hybrid representations are intrinsically tied to $\theta$, maintaining strict on-policy training. This makes M3PO both lightweight and compatible with other RL optimizations.

\paragraph{Inference.} 
M3PO employs single-path decoding at test time, matching standard LLMs inference. This efficiency is enabled by the reinforcement learning objective, which trains the policy $\pi_\theta$ to maximize the reward via a multi-path collaboration mechanism. As the policy iteratively updates to capture high-reward trajectories, it effectively internalizes robust reasoning patterns and learns a more reliable and self-consistent way of thinking. Thus, even in the absence of explicit path interaction at test time, the optimized policy still exhibits enhanced reasoning accuracy and stability.

\section{Experiments}
\label{sec:experiments}
We evaluate M3PO on two benchmark categories. The first is knowledge-intensive tasks, including open-domain and multi-hop question answering (QA), which assess factual utilization and evidence aggregation; the second is reasoning-intensive STEM problems that probe structured problem solving in scientific and mathematical settings.
\begin{table}[t]
\caption{Evaluation performance on QA benchmarks. This table reports exact match scores on five open-domain and multi-hop QA datasets using top-3 retrieved documents. The upper section shows RAG baselines with Qwen2.5-7B, while the lower sections show the performance of smaller Qwen models (1.5B and 3B) under different training strategies. M3PO delivers consistently strong performance across all datasets, with especially notable gains on the NQ dataset, demonstrating its effectiveness in enhancing reasoning robustness through internal refinement of reasoning processes.}
\label{tab:QA_results}
\tablestyle{1.8pt}{1.2}
\begin{tabular}{ccccccc}
\shline
           & NQ    & TriviaQA & HotpotQA & 2WikiMQA & Bamboogle & Average \\ \shline
\multicolumn{7}{c}{Qwen2.5-7B-Instruct}                                   \\ \shline
QA         & 13.4 & 40.8    & 18.3    & \textbf{25.0}    & 12.0      & 21.9   \\
CoT        & 4.8 & 18.5    & 9.2    & 11.1    & 23.2     & 13.4   \\
IRCoT      & 22.4 & 47.8    & 13.3    & 14.9    & 22.4     & 24.2   \\
Search-o1  & 15.1 & 44.3    & 18.7    & 17.6    & \textbf{29.6}     & 25.1   \\
RAG        & \textbf{34.9} & \textbf{58.5}    & \textbf{29.9}    & 23.5    & 20.8     & \textbf{33.5}   \\ \shline
\multicolumn{7}{c}{Qwen2.5-1.5B-Instruct}                                 \\ \shline
SFT        & 9.4 & 19.3    & 12.9    & 21.0     & 2.4     & 13.0    \\
RAG        & 28.8 & 47.7    & 22.8    & 20.3    & 7.2     & 25.4   \\
PPO        & 32.7 & 52.7    & 25.6    & 24.2    & 18.4     & 30.7   \\
GRPO       & 29.3 & 48.0     & 20.2    & 21.3    & 12.0      & 26.1   \\
HRPO       & 36.4 & 55.3    & 27.3    & 27.6    & 21.6     & 33.7   \\
\rowcolor[HTML]{EFF6FB} 
M3PO &  \textbf{41.4}     &    \textbf{56.8}      &  \textbf{28.7}        &   \textbf{27.9}       &   \textbf{23.2}        &     \textbf{35.6}    \\ \shline
\multicolumn{7}{c}{Qwen2.5-3B-Instruct}                                   \\ \shline
SFT        & 24.9 & 29.2    & 18.6    & 24.8    & 11.2     & 21.7   \\
RAG        & 34.8 & 54.4    & 25.5    & 22.6    & 8.0      & 29.1   \\
PPO        & 35.6 & 56.3    & 30.4    & 29.3    & 24.0      & 35.1   \\
GRPO       & 38.1 & 57.0     & 30.8    & 30.3    & 27.2     & 36.7   \\
HRPO       & 37.8 & 59.3    & 31.6   & \textbf{31.8}   & 29.6     & 38.0  \\
\rowcolor[HTML]{EFF6FB} 
M3PO &  \textbf{44.1}     &  \textbf{61.0}        &    \textbf{33.2}      &    31.4      &       \textbf{31.2 }   &   \textbf{40.2} \\
% \hline
\end{tabular}
\vspace{-10pt}
\end{table}

\subsection{Evaluation on Knowledge Benchmarks}
\paragraph{Datasets and baselines.} For knowledge-intensive reasoning, we use five widely adopted open-domain and multi-hop QA datasets: Natural Questions (NQ)~\cite{kwiatkowski2019natural}, TriviaQA~\cite{joshi2017triviaqa}, HotpotQA~\cite{yang2018hotpotqa}, 2WikiMultiHopQA (2WikiMQA)~\cite{ho2020constructing}, and Bamboogle~\cite{press2022measuring}. Following HRPO~\cite{yue2025hybrid}, we employ the E5-base embedding model~\cite{wang2022text} to retrieve the top three Wikipedia documents as context for each prompt. We train on a joint corpus formed by merging the NQ and HotpotQA training sets, and we report exact match results on each dataset’s official evaluation split. Detailed implementation settings are provided in the Appendix.

For the 1.5B and 3B model sizes~\cite{qwen2.5techreport}, we evaluate M3PO against representative post-training and retrieval baselines, including Supervised Fine-Tuning (SFT), the RL methods PPO~\cite{schulman2017proximal} and GRPO~\cite{shao2024deepseekmath}, Retrieval-Augmented Generation (RAG)~\cite{lewis2020retrieval}, and Hybrid Reasoning Policy Optimization (HRPO)~\cite{yue2025hybrid}, an RL-based approach built on soft thinking~\cite{zhang2025soft}. We also report results with the larger Qwen2.5-7B-Instruct~\cite{qwen2.5techreport} backbone using standard QA and retrieval pipelines, namely direct inference (QA),  Chain-of-Thought (CoT)~\cite{wei2022chain}, interleaving retrieval with CoT (IRCoT)~\cite{trivedi2023interleaving}, Search-o1~\cite{li2025search}, and RAG~\cite{lewis2020retrieval}. The best result in each block of Table \ref{tab:QA_results} is highlighted in bold.

\paragraph{Results analysis.}
Table~\ref{tab:QA_results} shows that M3PO consistently achieves the highest average exact match (EM) scores on knowledge-intensive benchmarks for both Qwen-1.5B and Qwen-3B. With Qwen-1.5B, M3PO attains 35.6\% EM, surpassing the strongest Qwen-7B RAG baseline by 2.1\%. With Qwen-3B, it reaches 40.2\% EM, a gain of 6.7\% over the same 7B baseline. These results demonstrate that training-time multi-path collaboration can effectively compensate for limited model capacity.

Relative to RL-based methods, M3PO yields substantial gains. In the 1.5B setting, it exceeds GRPO by 9.5\%, which supports the claim that explicit cross-path collaboration improves policy learning beyond group-relative updates alone. The advantage is particularly pronounced on the NQ dataset, where M3PO reaches 41.4\% EM at 1.5B and 44.1\% at 3B, representing improvements of 6.5\% and 9.2\% over the 7B RAG baseline. Moreover, M3PO consistently outperforms HRPO, another RL-based hybrid reasoning method, achieving higher average scores while remaining parameter-free. This further confirms M3PO's advantage in enabling lightweight and efficient deployment.

These results validate that by leveraging parallel rollouts as independent reasoning sources and enabling cross-path collaboration, M3PO cultivates more robust reasoning patterns without sacrificing efficiency or architectural changes.

\subsection{Evaluation on STEM Benchmarks}

\paragraph{Datasets and baselines.} 
For reasoning-intensive STEM evaluation, we employ five benchmarks spanning mathematical and scientific domains: GSM8k~\cite{cobbe2021training}, MATH~\cite{hendrycks2021measuring}, MATH500~\cite{lightman2023let}, MMLU-STEM~\cite{hendrycks2020measuring}, and ARC-Challenge~\cite{clark2018think}. For training, we use the GSM8k training split for GSM8k, the MATH training split for MATH and MATH500, and a merged corpus that combines the auxiliary MMLU and ARC-C training sets for MMLU-ST and ARC-C.
Detailed settings are provided in Appendix.

% For reasoning-intensive STEM evaluation, we employ five benchmarks spanning mathematical and scientific domains: GSM8k~\cite{cobbe2021training} (grade-school arithmetic problems), MATH~\cite{hendrycks2021measuring} (diverse competition-level mathematics), MATH500~\cite{lightman2023let} (a curated subset of MATH), MMLU-STEM~\cite{hendrycks2020measuring} (MMLU-ST; science and engineering multiple-choice questions), and ARC-Challenge~\cite{clark2018think} (ARC-C; complex science reasoning). For training, we use the GSM8k training split for GSM8k, the MATH training split for MATH and MATH500, and a merged corpus that combines the auxiliary MMLU and ARC-C training sets for MMLU-ST and ARC-C. Detailed implementation settings are provided in Appendix A.

For models at the 1.5B and 3B scales~\cite{qwen2.5techreport}, we compare against several strong baselines: SFT, standard  RL methods like PPO~\cite{schulman2017proximal} and GRPO~\cite{shao2024deepseekmath}, and the RL-based latent reasoning method HRPO~\cite{yue2025hybrid}. To further contextualize the performance of M3PO, we also include comparisons with larger models ($\geq$ 7B parameters) using few-shot CoT reasoning, specifically DeepSeekMath-7B~\cite{shao2024deepseekmath}, Gemma-2-9B~\cite{team2024gemma}, Qwen2.5-7B~\cite{qwen2.5techreport}, and MAmmoTH2-7B~\cite{yue2024mammoth2}. 

% For models at the 1.5B and 3B scales~\cite{qwen2.5techreport}, we compare against several strong baselines: SFT, SFT with distilled CoT~\cite{team2025qwq} (Distill CoT), standard  RL methods like PPO~\cite{schulman2017proximal} and GRPO~\cite{shao2024deepseekmath}, as well as the RL-based latent reasoning method HRPO~\cite{yue2025hybrid}. To further contextualize the performance of M3PO, we additionally include comparisons with larger models ($\geq$ 7B parameters) using few-shot CoT reasoning, specifically DeepSeekMath-7B~\cite{shao2024deepseekmath}, Gemma-2-9B~\cite{team2024gemma}, Qwen2.5-7B~\cite{qwen2.5techreport}, and MAmmoTH2-7B~\cite{yue2024mammoth2}. 

\paragraph{Results analysis.}
% Table \ref{tab:stem_results} shows that M3PO delivers the strongest results with compact Qwen backbones and can match the performance of much larger language models. Several observations emerge. First, SFT underperforms relative to RL methods, indicating the effectiveness of RL with verifiable rewards on reasoning-intensive STEM tasks. Second, with the 3B backbone, M3PO attains an average accuracy of 70.5\%, outperforming the 7B leader baseline by 5.3\%. Third, compared with the hybrid reasoning approach HRPO, M3PO achieves stronger results on the more challenging tasks, suggesting that collaborative learning helps internalize more robust reasoning patterns. Fourth, M3PO sets the highest reported accuracy for sub-7B models on MATH (0.613) and surpasses the leading 7B model on MATH by 10.9\%, highlighting the value of RL-based collaborative reasoning on difficult benchmarks.
As shown in Table \ref{tab:stem_results}, M3PO achieves state-of-the-art performance with compact Qwen models while competing effectively with substantially larger LLMs. 

In particular, SFT consistently underperforms RL-based approaches, confirming the importance of verifiable rewards for complex reasoning. Moreover, M3PO with a 3B model attains an average accuracy of 70.5\%, outperforming the strongest 7B baseline by 5.3\%. It also surpasses HRPO on all datasets, indicating that its collaborative learning mechanism better internalizes robust reasoning patterns. A compelling case is the MATH benchmark, where the 3B model trained with M3PO achieves 60.7\% accuracy, significantly exceeding the best 7B baseline by 10.9\%.

Collectively, these findings validate that M3PO's collaborative reasoning paradigm effectively unlocks large-model capabilities within compact architectures, demonstrating a promising direction for developing an efficient yet powerful reasoning framework.

\begin{table}[t]
\caption{Evaluation performance on STEM benchmarks. This table reports accuracy on five reasoning-intensive datasets. The upper section shows few-shot results from LLMs with at least 7B parameters, while the lower sections present the performance of smaller Qwen models (1.5B and 3B) under various training strategies. M3PO achieves the best or competitive performance across all datasets, demonstrating that the multi-path collaborative mechanism effectively enhances complex reasoning and supports strong generalization via policy optimization.}
\label{tab:stem_results}
\tablestyle{0.7pt}{1.2}
\begin{tabular}{ccccccc}
\shline
 & GSM8k & MATH & MATH500 & MMLU-ST & ARC-C & Average \\ \shline
\multicolumn{7}{c}{Larger LLMs (Size $\geq$ 7B)} \\ \shline
DeepSeekMath-7B & 64.2 & 36.2 & 34.6 & 56.5 & 67.8 & 51.9 \\
Gemma-2-9B & 70.7 & 37.7 & 36.4 & 65.1 & 68.2 & 55.6 \\
Qwen2.5-7B & \textbf{85.4} & \textbf{49.8} & 46.4 & \textbf{72.3} & 63.7 & 63.5 \\
MAmmoTH2-7B & 68.4 & 36.7 & 39.6 & 62.4 & 81.7 & 57.8 \\ 
MAmmoTH2-8B & 70.4 & 35.8 & \textbf{73.2} & 64.2& \textbf{82.2} & \textbf{65.2}\\ \shline
\multicolumn{7}{c}{Qwen2.5-1.5B-Instruct} \\ \shline
SFT & 56.0 & 30.0 & 30.2 & 40.3 & 60.2 & 43.3 \\
% Distilled CoT & 70.6 & 50.3 & - & - & - & - \\
PPO & 67.6 & 45.4 & 44.8 & 56.6 & 71.5 & 57.2 \\
GRPO & 68.2  & 46.0  & 45.2  & 56.2 & 73.7 & 57.9 \\
HRPO & 69.9  & 43.8  & 45.8  & 56.9 & 74.2 & 58.1 \\
\rowcolor[HTML]{EFF6FB} 
M3PO & \textbf{70.2} & \textbf{47.4} & \textbf{48.0} & \textbf{58.1} & \textbf{77.6} & \textbf{60.3} \\ \shline
\multicolumn{7}{c}{Qwen2.5-3B-Instruct} \\ \shline
SFT & 67.0 & 34.8 & 36.0 & 45.4 & 47.4 & 46.1 \\
% Distilled CoT & 79.9 & 57.5 & - & - & - & - \\
PPO & 81.9 & 59.7 & 60.4 & 58.2 & 81.1 & 68.2 \\
GRPO & 83.4 & 60.2 & 60.4 & 60.1 & 81.4 & 69.1 \\
HRPO & 83.5 & 58.6 & 60.2 & 59.0 & 82.0 & 68.7 \\
\rowcolor[HTML]{EFF6FB} 
M3PO & \textbf{84.8} & \textbf{60.7} & \textbf{63.0} & \textbf{61.6} & \textbf{82.6} & \textbf{70.5} \\
% \hline
\end{tabular}
\vspace{-10pt}
\end{table}

\subsection{Ablation Study} %
\paragraph{Latent reasoning paradigm.} We integrate different latent reasoning paradigms into the same policy optimization framework and train the Qwen2.5-3B-Instruct model on the MATH dataset to ensure fair comparison. The variants are: (1) Hidden States, which feed the final-layer hidden state back as the next input; (2) Soft Thinking, which forms the next input by a soft weighted aggregation in the input embedding space; (3) HRPO, a hybrid approach that combines the sampled next input with a soft aggregation via a gating mechanism; and (4) M3PO, our hybrid reasoning method with explicit cross-path interaction (see Equation (\ref{eq:hybrid})).

\begin{figure}[t!]
    \centering
    \includegraphics[width=1.0\linewidth]{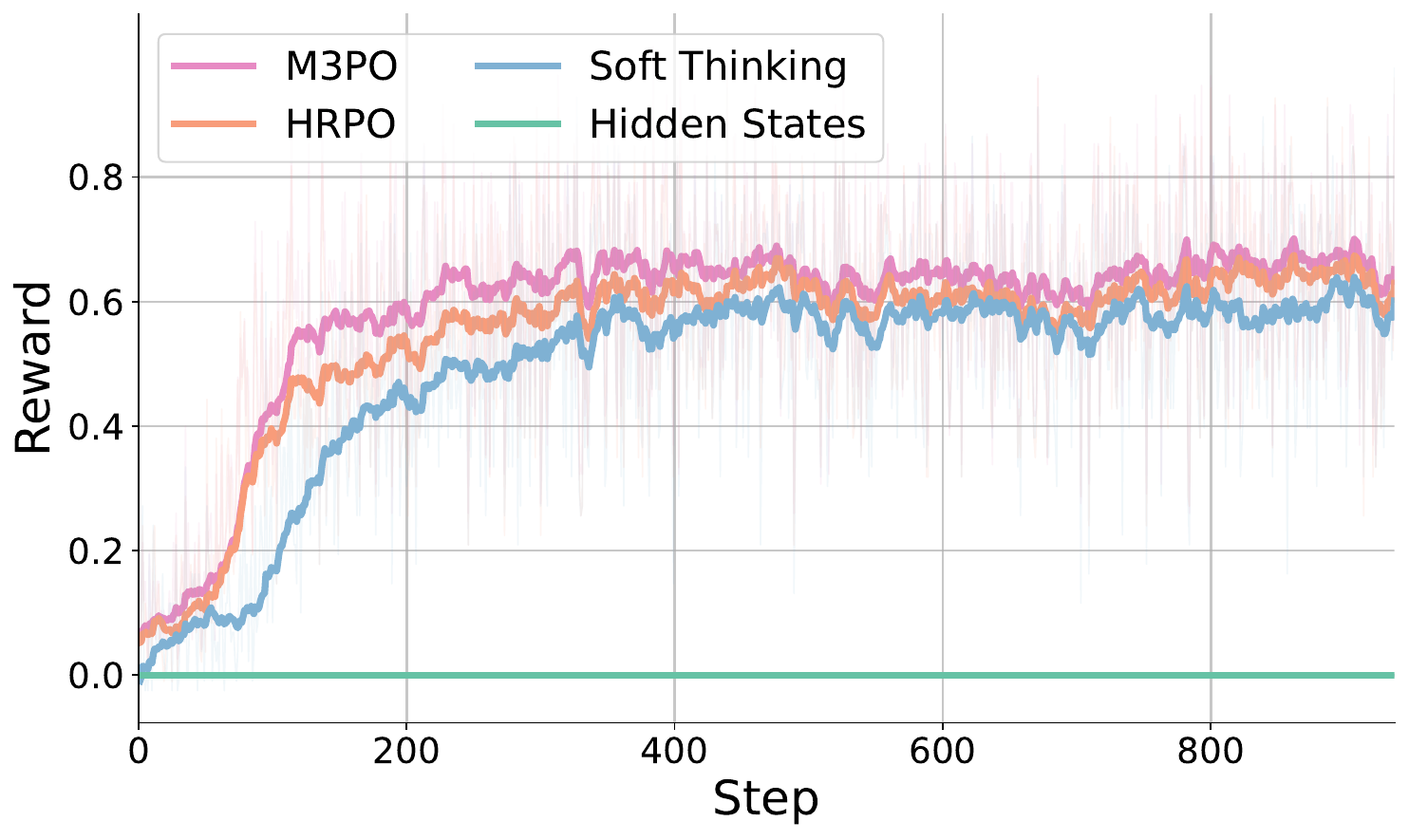}
    \caption{Training performance across latent reasoning variants. M3PO not only converges more rapidly but also attains consistently higher final rewards, underscoring the efficacy and robustness of our hybrid reasoning paradigm in complex reasoning tasks.
    }
    \label{fig:ablate_latent_reasoning}
    \vspace{-10pt}
\end{figure}

Figure \ref{fig:ablate_latent_reasoning} visualizes the exponential moving average (EMA) of training rewards. The hidden states approach remains zero reward, primarily due to the distributional discrepancy between hidden states and the pretrained embedding space, leading to incompatibility and performance degradation. 
While both Soft Thinking and HRPO utilize the soft aggregation scheme, Soft Thinking exhibits slower convergence and attains lower reward levels. This performance pattern highlights the benefits of hybrid reasoning and underscores the importance of preserving the distinctive characteristics of each reasoning trajectory. 

In contrast, M3PO achieves both faster convergence and higher stabilized reward levels compared to HRPO, although both methods employ hybrid reasoning principles. This performance differential confirms the superior effectiveness of our explicit multi-path interaction mechanism over conventional soft aggregation methods, validating the efficacy of structured cross-path collaboration in leveraging independent rollouts. Importantly, unlike HRPO which requires additional parameters for its gating mechanism, M3PO maintains a fully parameter-efficient design while delivering stronger performance.

% While Soft Thinking initially performs comparably to M3PO in the first few hundred steps, its rewards eventually collapse and recover only slowly, likely because continuous soft aggregation introduces excessive noise over time. Although both M3PO and HRPO adopt hybrid reasoning, M3PO achieves superior training rewads without introducing additional parameters, demonstrating the efficacy of our collaborative mechanism that effectively leverages independent rollouts through explicit cross-path interaction.

\begin{figure}[t!]
    \centering
    \includegraphics[width=1.0\linewidth]{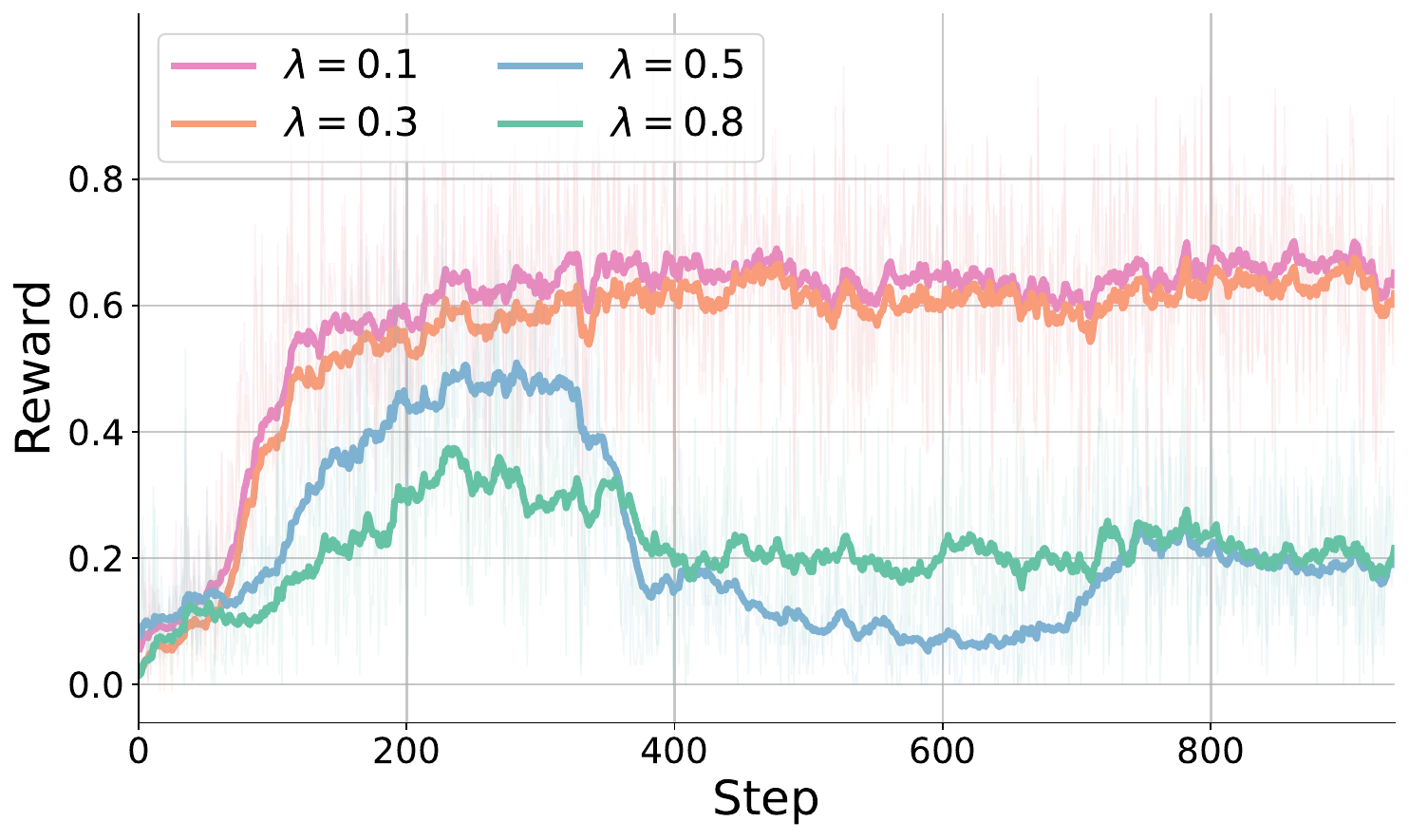}
    \caption{Sensitivity of the blending coefficient $\lambda$ in Equation~(\ref{eq:hybrid}). Performance collapses when $\lambda \ge 0.5$, highlighting the critical balance between intrinsic reasoning direction and peer insights.}
    \label{subfig:lambda_reward}
    \vspace{-10pt}
\end{figure}

% baseline: HRPO, SOFT THINKING, HIDDEN STATES
\paragraph{Effect of cross-path fusion mechanism.} 
To evaluate the impact of our cross-path fusion strategy based on distributional similarity, we compare against two alternative approaches: (1) Peer Mean, which averages peer embeddings without gating; (2) No Cross-Path, which entirely removes peer insights and operates like conventional CoT reasoning.

\begin{figure}[t!]
    \centering
    \includegraphics[width=1.0\linewidth]{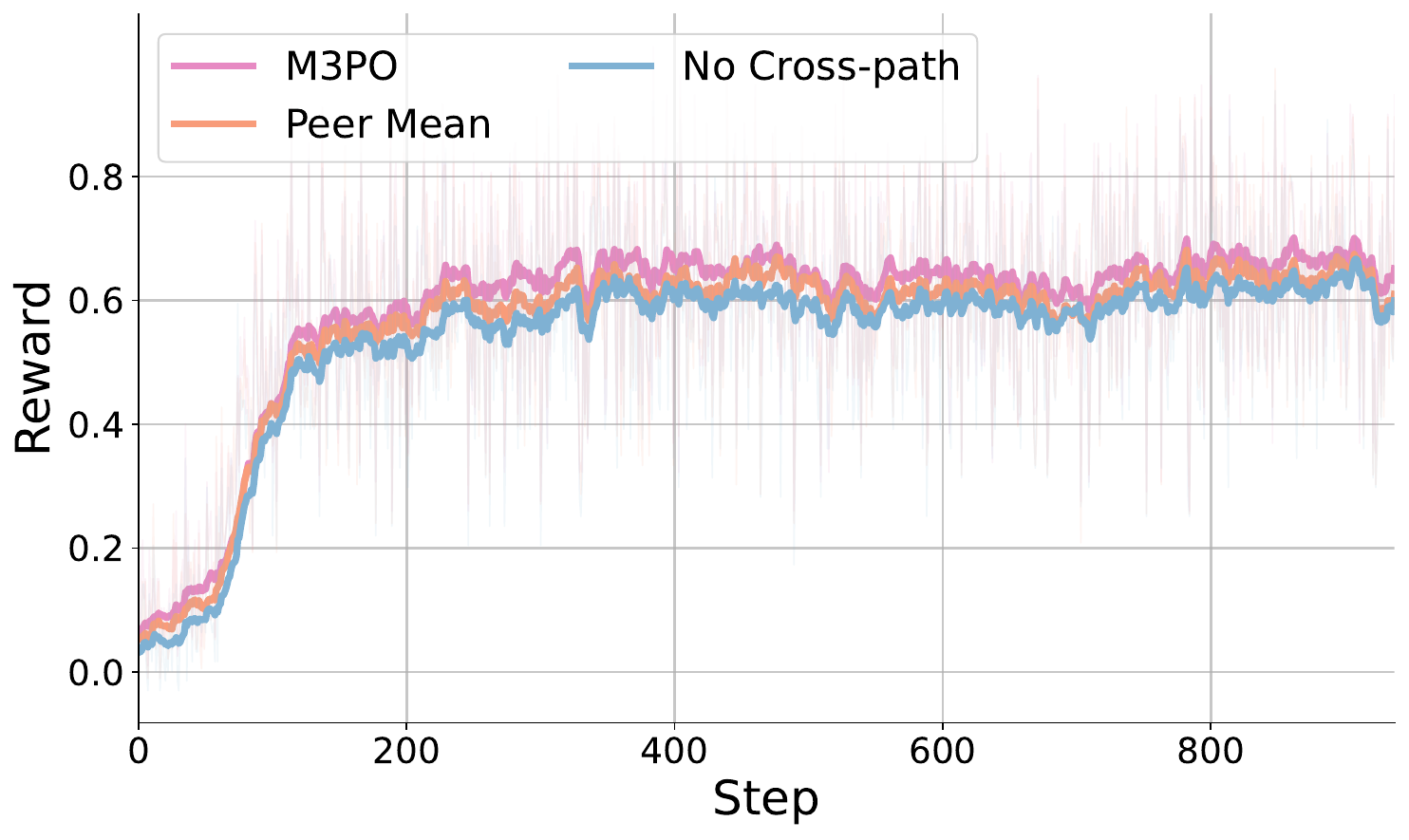}
    \caption{Ablation study on the impact of the distribution-similarity fusion mechanism. M3PO achieves the highest rewards, highlighting the effectiveness of our warm cross-path fusion strategy and the value of peer insights.
    }
    \label{fig:ablate_fusion}
    \vspace{-10pt}
\end{figure}

Figure \ref{fig:ablate_fusion} shows the EMA of training rewards on MATH with Qwen2.5-3B. Replacing M3PO's similarity-guided fusion with uniform averaging, which removes selective gating and assigns equal weight to all peers, leads to measurable performance degradation. This result confirms that distributional divergences among trajectories introduce conflicting signals, hindering effective utilization of complementary reasoning paths. The performance degradation stems from a fundamental incompatibility between uniform aggregation and the structured reasoning processes inherent in pretrained models. This mismatch introduces propagating noise that progressively undermines both reasoning coherence and learning stability.

% This result confirms that distributional divergences among trajectories could introduce conflicting signals, which hinders the effective exploitation of complementary reasoning paths. The performance decline stems from a fundamental mismatch between uniform aggregation and the structured reasoning processes expected by the pretrained model. This mismatch introduces noise that accumulates throughout the reasoning trajectory, ultimately compromising both coherence and learning stability.

\begin{figure}[t!]
    \centering
    \includegraphics[width=1.0\linewidth]{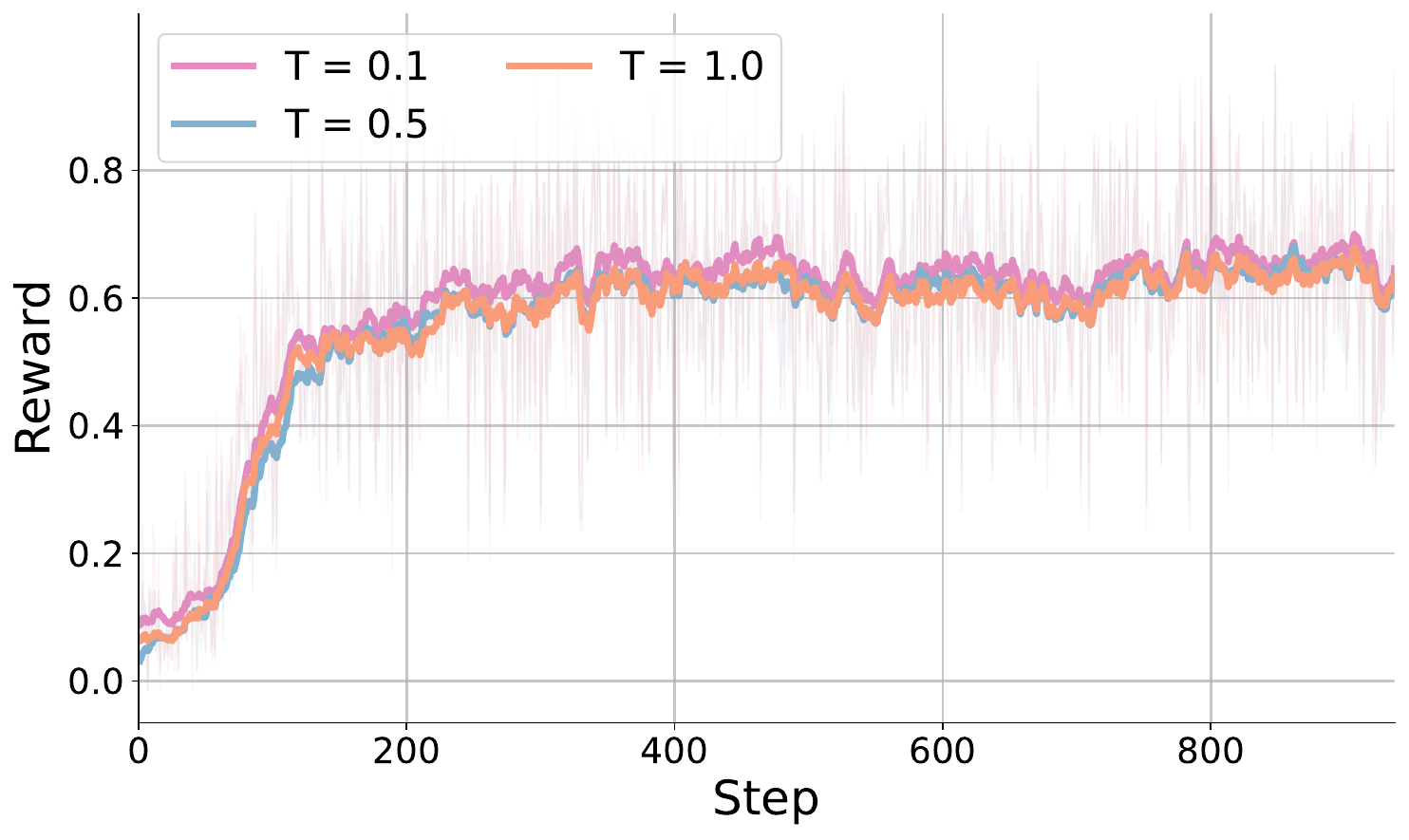}
    \caption{
    Sensitivity of the temperature $T$ in Equation~(\ref{eq:temp_softmax}). Performance peaks at $T=0.1$, where sharp attention weights enhance reasoning by selectively focusing on relevant peers. 
}
    \label{fig:ablate_T}
    \vspace{-10pt}
\end{figure}

\begin{figure*}[h]
    \centering
    \includegraphics[width=1.0\linewidth]{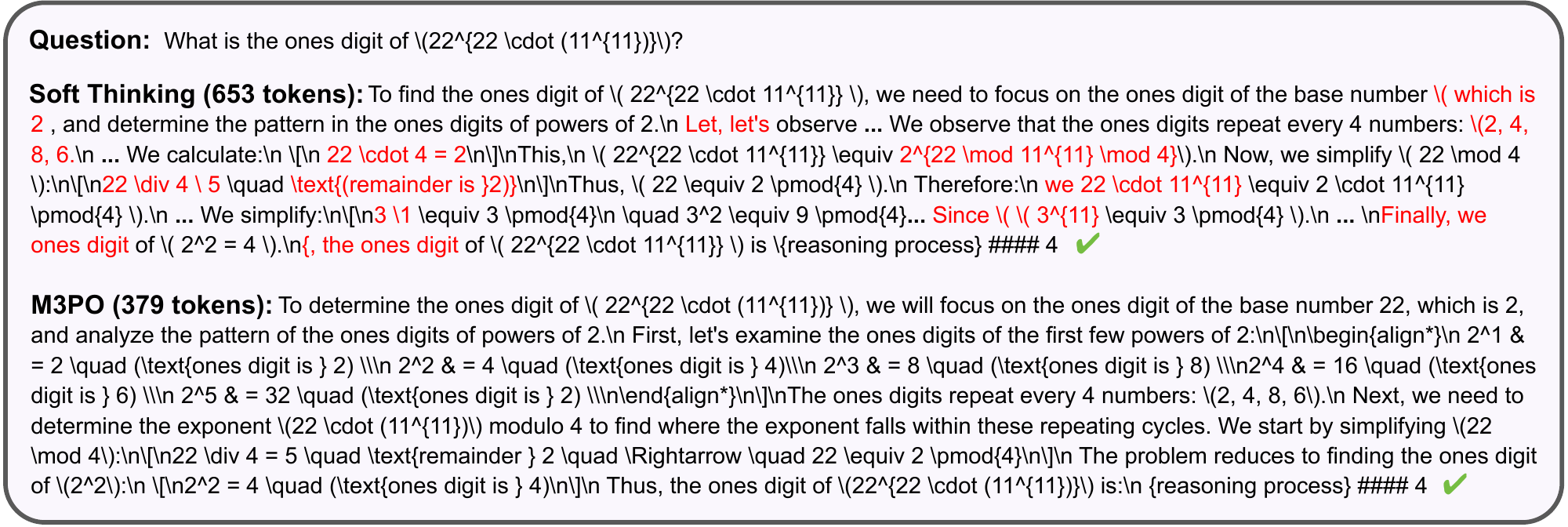}
    \caption{A Comparison of training rollouts for Soft Thinking and M3PO on a mathematical problem. The complete Soft Thinking rollout is provided in the Appendix. Text highlighted in \textcolor{red}{red} indicates erroneous insertions, discontinuities, or formatting defects.  Soft Thinking exhibits noisy and spurious text, while M3PO exhibits clean steps with consistent formatting.}
    \label{fig:case_study}
    \vspace{-10pt}
\end{figure*}

Furthermore, complete removal of peer insights yields the lowest performance when the model reverts to standard CoT reasoning. This ablation result consistently underperforms both M3PO and the peer mean variant, strongly validating two crucial aspects of our approach. First, incorporating multi-path perspectives provides fundamental benefits over isolated reasoning trajectories. Second, our gated collaboration mechanism offers specific advantages over naive aggregation methods by selectively integrating the most consistent cross-path signals, thereby enhancing both reasoning robustness and training stability.
% AVERAGE, ENTROPY

\paragraph{Sensitivity of $\lambda$ on hybrid reasoning.}
M3PO is built on the hybrid reasoning scheme, which introduces a blending coefficient $\lambda$ in Equation~(\ref{eq:hybrid}). To systematically examine the effect of this balancing strategy, we conduct sensitivity analysis with $\lambda$ values from [0.1, 0.3, 0.5, 0.8].

Figure \ref{subfig:lambda_reward} illustrates the training rewards on the MATH dataset using Qwen2.5-3B. A clear performance collapse occurs when $\lambda\geq0.5$, where the anchor trajectory loses its dominant role in the hybrid fusion. This outcome aligns with expectations, as substantially diminishing the contribution of the anchor trajectory disrupts the logical coherence essential for valid multi-step inference. Such disruption proves particularly critical in autoregressive models, where local inconsistencies propagate and amplify through subsequent reasoning steps. 
% Figure~7 illustrates representative case studies. At $\lambda=0.8$, the reasoning shows semantic fragmentation and unstable token sequences, while $\lambda=0.1$ maintains coherent and logically consistent chains.

Moreover, $\lambda=0.1$ achieves consistently better performance than $\lambda=0.3$, while Figure \ref{fig:ablate_fusion} confirms that $\lambda=0.1$  (M3PO) outperforms the $\lambda=0$  ablation (No Cross-Path). This indicates that optimal performance requires both preserving the structural integrity of the original reasoning flow and allowing controlled integration of peer-derived insights. Striking this balance allows the model to harness complementary reasoning signals without compromising coherence, maximizing its overall problem-solving capability.

% \begin{figure*}[t!]
%     \centering
%     \captionsetup[subfigure]{margin=2pt}
%     % 设置子标题编号的位置是120pt
%     \subfloat[Reward.]{
%     \label{subfig:lambda_reward}
%         \centering
%     \includegraphics[width=0.45\linewidth]{figures/lambda_reward.pdf}
%     }
%     \hspace{15pt}
%     \subfloat[Completion length.]{
%     \label{subfig:lambda_length}
%         \centering
%     \includegraphics[width=0.45\linewidth]{figures/lambda_length.pdf}
%     }
%     \caption{Sensitivity analysis of bending parameter $\lambda$ in Equation (\ref{eq:hybrid}). Training reward and completion length are shown for different values of $\lambda \in [0.1, 0.3, 0.5, 0.8]$. A sharp performance collapse occurs when $\lambda \geq 0.5$, underscoring the critical importance of maintaining distinctive reasoning trajectories through balanced hybrid thinking.}
%     \label{fig:vim_vs_ours}
% \end{figure*}

% \begin{figure}
%     \centering
%     \includegraphics[width=1.0\linewidth]{figures/lambda_reward.pdf}
%     \caption{Caption}
%     \label{fig:placeholder}
% \end{figure}
% \begin{figure}
%     \centering
%     \includegraphics[width=1.0\linewidth]{figures/lambda_length.pdf}
%     \caption{Caption}
%     \label{fig:placeholder}
% \end{figure}
% tau=0.1: 0.1, 0.3, 0.5,0.8
% tau=1.0: 0.1, 0.3, 0.5, 0.8

\paragraph{Sensitivity of $T$ on collaborative learning.}
To regulate information flow in multi-path collaborative learning, we introduce a temperature parameter $T$ that governs the selectivity of attention over peer trajectories. We assess its effectiveness via sensitivity analysis with $T \in [0.1, 0.5, 1.0]$.

Figure \ref{fig:ablate_T} illustrates the training rewards on the MATH dataset using Qwen2.5-3B. The results show that $T=0.1$ achieves superior performance, while both $T=0.5$ and $T=1.0$ exhibit comparable but lower performance levels. This confirms that sharper attention weights enhance collaboration quality by concentrating on the most relevant cross-path signals, whereas higher temperatures reduce selectivity and compromise reasoning stability through overly dispersed attention distributions.

The temperature parameter $T$ works synergistically with the blending parameter $\lambda$ to create a dual-control mechanism for robust multi-path reasoning. While $\lambda$ balances the influence between the intrinsic reasoning direction and external insights, the temperature $T$ fine-tunes the quality of incorporated external signals by filtering out less relevant contributions. This hierarchical gating strategy ensures that the model selectively perceives the most beneficial cross-path information while maintaining training stability and compatibility with the pretrained architecture.

\subsection{Qualitative Results}
% \paragraph{Corrected Patterns}
Figure \ref{fig:case_study} compares reasoning rollouts of Soft Thinking and M3PO under identical optimization conditions.  Soft Thinking produces erroneous insertions and logical discontinuities, suggesting that soft aggregation can propagate and amplify noise during autoregressive generation. In contrast, M3PO maintains a coherent, well-structured reasoning chain throughout the process, demonstrating its superior stability and noise resistance.  This qualitative difference underscores the value of structured cross-path collaboration in ensuring reasoning coherence and interpretability.

\section{Conclusion}
\label{sec:conclusion}
In this work, we present M3PO, a multi-path collaborative RL framework that enhances reasoning robustness through structured trajectory interaction. By integrating cross-path insights into policy optimization, M3PO develops more reliable reasoning patterns while maintaining training stability. Extensive experiments demonstrate consistent improvements across various benchmarks.\\
\textbf{Limitations.} Computational resources limited our exploration to models up to 3B parameters. Future work will investigate M3PO's scalability and adaptive collaboration mechanisms. Despite this, M3PO establishes a solid foundation for multi-path collaborative reasoning that effectively balances performance and deployment feasibility.

% Our experiments are limited to models up to 3B parameters due to computational constraints. Future work will explore M3PO’s scalability and adaptive collaboration mechanisms. Despite this, our framework offers a practical foundation for multi-path reasoning that balances performance and deployment feasibility.

{
    \small
    \bibliographystyle{ieeenat_fullname}
    \bibliography{main}
}

\clearpage
\setcounter{page}{1}
\maketitlesupplementary

\section{Implementation}
\label{sec:implementation}
M3PO is a lightweight multi-path collaborative learning framework designed to be compatible with any LLM architecture. It introduces no additional trainable parameters and preserves the original inference efficiency of the base model. To enable efficient training under this framework, we integrate optimized kernel implementations from Unsloth\footnote{\url{https://github.com/unslothai/unsloth}} and apply low-rank adaptation (LoRA)~\cite{hu2022lora} for parameter-efficient fine-tuning. All M3PO experiments adopt the hyperparameter settings detailed in Table~\ref{tab:hyperparameters}, which serve as our default unless otherwise specified.

\begin{table}[htbp]
\centering
\caption{Training settings for M3PO.}
\label{tab:hyperparameters}
\begin{tabular}{ll}
\toprule
Algorithm & M3PO \\
Epochs & 1 \\
Optimizer & AdamW 8bit \\
Optimizer Momentum & $\beta_1$, $\beta_2 = 0.9$, $0.99$ \\
Weight Decay & 0.1 \\
Learning Rate & 5e-6 \\
M3PO $\beta$ & 0.005 \\
Max Gradient Norm & 0.1 \\
Gradient Accumulation Step & 4 \\
Group size $N$ in M3PO & 4 / 8 \\
Total Train Batch Size & 64 \\
LR Scheduler & Cosine with Warmup \\
Warmup Ratio & 0.1 \\
Precision (WA) & BF16-mixed \\
Completion Length & 1024 \\
$\lambda$ in Equation~(\ref{eq:hybrid}) & 0.1\\
$T$ in Equation~(\ref{eq:temp_softmax}) & 0.1\\
\midrule
LoRA Modules & query, key, value, dense \\
LoRA Rank & 32 \\
LoRA $\alpha$ & 64 \\
\bottomrule
\end{tabular}
\end{table}

Thanks to the lightweight design of M3PO and our optimized kernels, M3PO runs efficiently on a single GPU for all tasks. Notably, all STEM benchmark datasets are used directly from their HuggingFace\footnote{\url{https://huggingface.co}} repositories without additional preprocessing.
We use a fixed group size of 4 for knowledge-intensive tasks. For complex reasoning benchmarks, including GSM8k, MATH, and MMLU-STEM, we generate 8 hybrid completions per query to enhance exploration and robustness.

The prompt construction follows a consistent template during both training and evaluation.  Each input begins with a system message instructing the model to carry out step-by-step reasoning before generating its final answer, followed by the user query. The full prompt is then formatted according to the model’s native chat template. In line with established practice~\cite{yue2025hybrid}, we use the minimal delimiter $\text{\#\#\#\#}$ to separate the model's reasoning chain from its final answer. This delimiter tokenizes as a single unit, adding no sequence length overhead while providing a clear signal for transitioning from latent collaborative reasoning to autoregressive answer generation. To ensure reasoning integrity, a penalty mechanism assigns zero reward to completions containing repeated delimiters, regardless of answer correctness. This prevents early termination of the reasoning process and encourages complete reasoning chains. Full prompt examples for different task types, including system messages and representative queries, are provided in Figures~\ref{fig:example_prompt_knowledge}, \ref{fig:example_prompt_math}, and \ref{fig:example_prompt_science}, respectively.

\begin{figure*}[htbp]
\centering
\begin{promptbox}[title={Example Prompt for Knowledge Tasks}]
<|im_start|>system
A conversation between User and Assistant. The user asks a question, and the assistant solves it. The assistant first thinks about the reasoning process in the mind and then provides the user with the answer. The final answer is provided after the #### tag, i.e., {reasoning process} #### {answer}.<|im_end|>
<|im_start|>user
Context (which may or may not be relevant):
Anton Zingarevich::::Anton Zingarevich Anton Zingarevich is a Russian businessman ...
Anton Zingarevich::::The couple married in late 2009 and had a child ...
Nigel Howe::::Nigel Howe Nigel Howe (born 7 April 1958) is a British property developer...

Question: who wrote the first declaration of human rights?<|im_end|>
<|im_start|>assistant
\end{promptbox}
\caption{Example prompt for knowledge tasks in M3PO. Context is partially omitted for brevity.}
\label{fig:example_prompt_knowledge}
\end{figure*}

We use greedy decoding and the standard inference pipeline of the base LLM to ensure reproducibility. For knowledge tasks, exact match scores are reported on validation and test splits following \cite{jin2025search}. For mathematical reasoning benchmarks including GSM8K, MATH, and MATH-500, along with multiple-choice datasets (MMLU-STEM and ARC-Challenge), we follow the post-processing and scoring procedures defined in~\cite{yue2024mammoth2}.

\begin{figure*}[tp]
\centering
\begin{promptbox}[title={Example Prompt for GSM8k / MATH / MATH500}]
<|im_start|>system 
A conversation between User and Assistant. The user asks a question, and the assistant solves it. The assistant first thinks about the reasoning process in the mind and then provides the user with the answer. The final answer is provided after the #### tag, i.e., {reasoning process} #### {answer}.<|im_end|> 
<|im_start|>user 
The square of an integer is 182 greater than the integer itself. What is the sum of all integers for which this is true?<|im_end|> 
<|im_start|>assistant
\end{promptbox}
\caption{Example prompt for GSM8k / MATH / MATH500 in M3PO.}
\label{fig:example_prompt_math}
\end{figure*}

\begin{figure*}[tp]
\centering
\begin{promptbox}[title={Example Prompt for MMLU-ST / ARC-C}]
<|im_start|>system 
A conversation between User and Assistant. The user asks a question, and the assistant solves it. The assistant first thinks about the reasoning process in the mind and then provides the user with the answer. The final answer is provided after the #### tag, i.e., {reasoning process} #### {answer}.<|im_end|> 
<|im_start|>user 
Question: Which of these do scientists offer as the most recent explanation as to why many plants and animals died out at the end of the Mesozoic era?  

Options: 
A. worldwide disease
B. global mountain building 
C. rise of mammals that preyed upon plants and animals
D. impact of an asteroid created dust that blocked the sunlight<|im_end|> 
<|im_start|>assistant
\end{promptbox}
\caption{Example prompt for MMLU-ST / ARC-C in M3PO.}
\label{fig:example_prompt_science}
\end{figure*}

\section{Additional Results}
\begin{table}[b]
\centering
\caption{Performance comparison of M3PO against alternative latent reasoning methods. M3PO achieves consistent improvements over all baselines on both benchmarks.}
\label{tab:latent_comp}
\tablestyle{0.8pt}{1.2}
\begin{tabular}{lcccccccc}
\toprule
 & \multicolumn{2}{c}{\textbf{Coconut}} & \multicolumn{2}{c}{\textbf{CODI}} & \multicolumn{2}{c}{\textbf{HRPO}} & \multicolumn{2}{c}{\textbf{M3PO}} \\
\cmidrule(lr){2-3}\cmidrule(lr){4-5}\cmidrule(lr){6-7}\cmidrule(lr){8-9}
 & GSM8k & MATH & GSM8k & MATH & GSM8k & MATH & GSM8k & MATH \\
\midrule
Accuracy & 31.5 & - & 65.8 & 41.9 & 69.9 & 43.8 & \textbf{70.2} & \textbf{47.4} \\
\bottomrule
\end{tabular}
\end{table}
\paragraph{Comparison to latent reasoning methods.} Beyond the RL baselines examined in our main experiments, we further compare M3PO against representative latent reasoning approaches, including HRPO, Coconut, and CODI. All methods are evaluated on the GSM8k and MATH benchmarks using the Qwen-1.5B backbone. For Coconut, we utilize its augmented CoT training data, while for CODI we adopt the original CoT trajectories from each dataset. HRPO is implemented under the same settings as M3PO.

As summarized in Table~\ref{tab:latent_comp}, M3PO achieves the highest accuracy on both reasoning tasks, outperforming all latent reasoning baselines by a consistent margin. Coconut lags notably on GSM8k, suggesting limitations in its token compression strategy for representing complex reasoning. Although CODI shows substantial gains from CoT fine-tuning, it still trails behind M3PO in final performance. HRPO, as a hybrid reasoning approach, shows competitive results but remains consistently behind M3PO across both benchmarks. These results collectively demonstrate that our approach maintains consistent advantages over existing latent reasoning methods, highlighting the effectiveness of explicit multi-path collaboration compared to implicit representation learning techniques.

\begin{figure*}[htbp]
    \centering
    \captionsetup[subfigure]{margin=2pt}
    % 设置子标题编号的位置是120pt
    \subfloat[Reward.]{
        \label{subfig:acc_plot}
        \centering
    \includegraphics[width=0.48\linewidth]{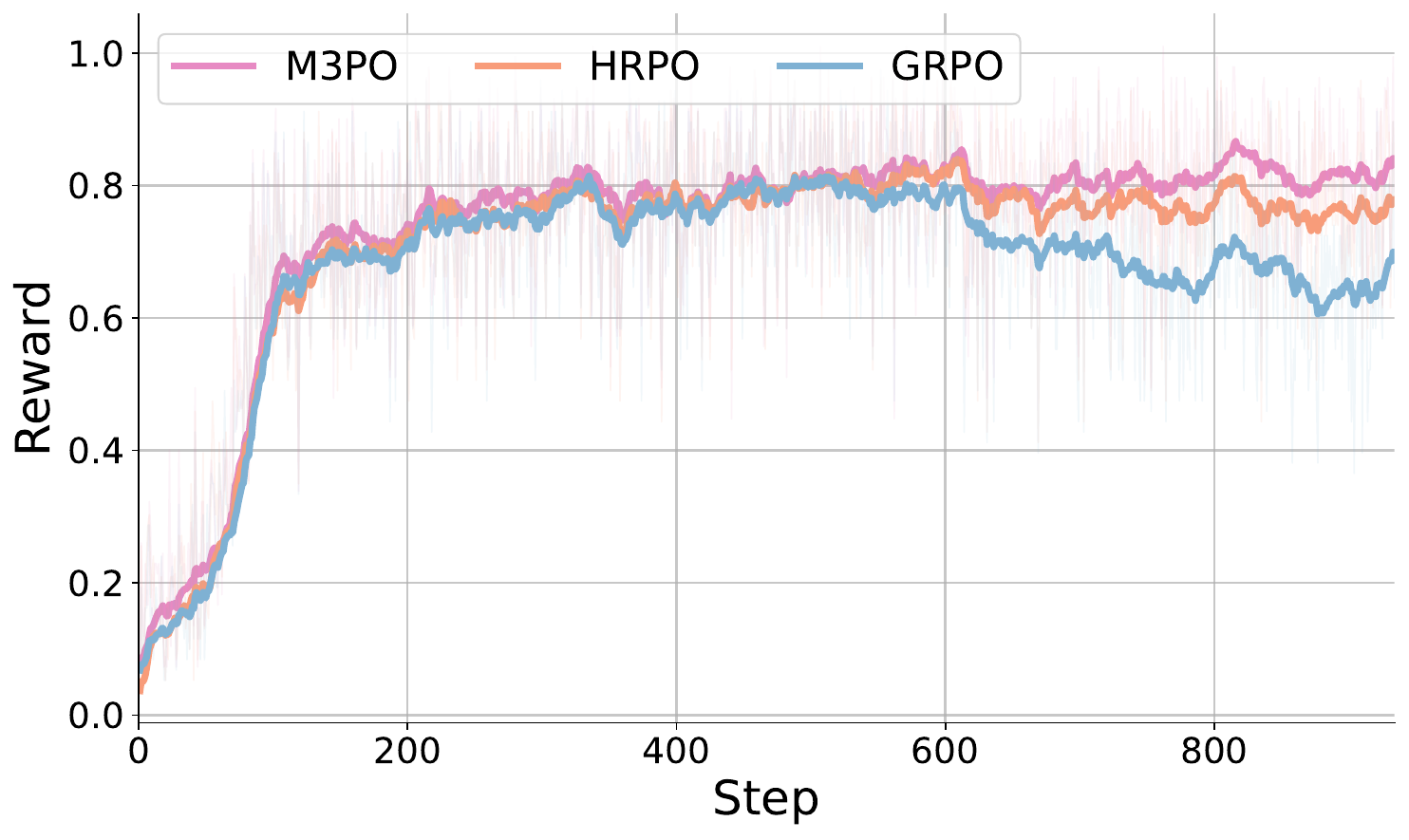}
    }
    \hspace{2pt}
    \subfloat[Completion length.]{
            \label{subfig:throughput_plot}
        \centering
    \includegraphics[width=0.48\linewidth]{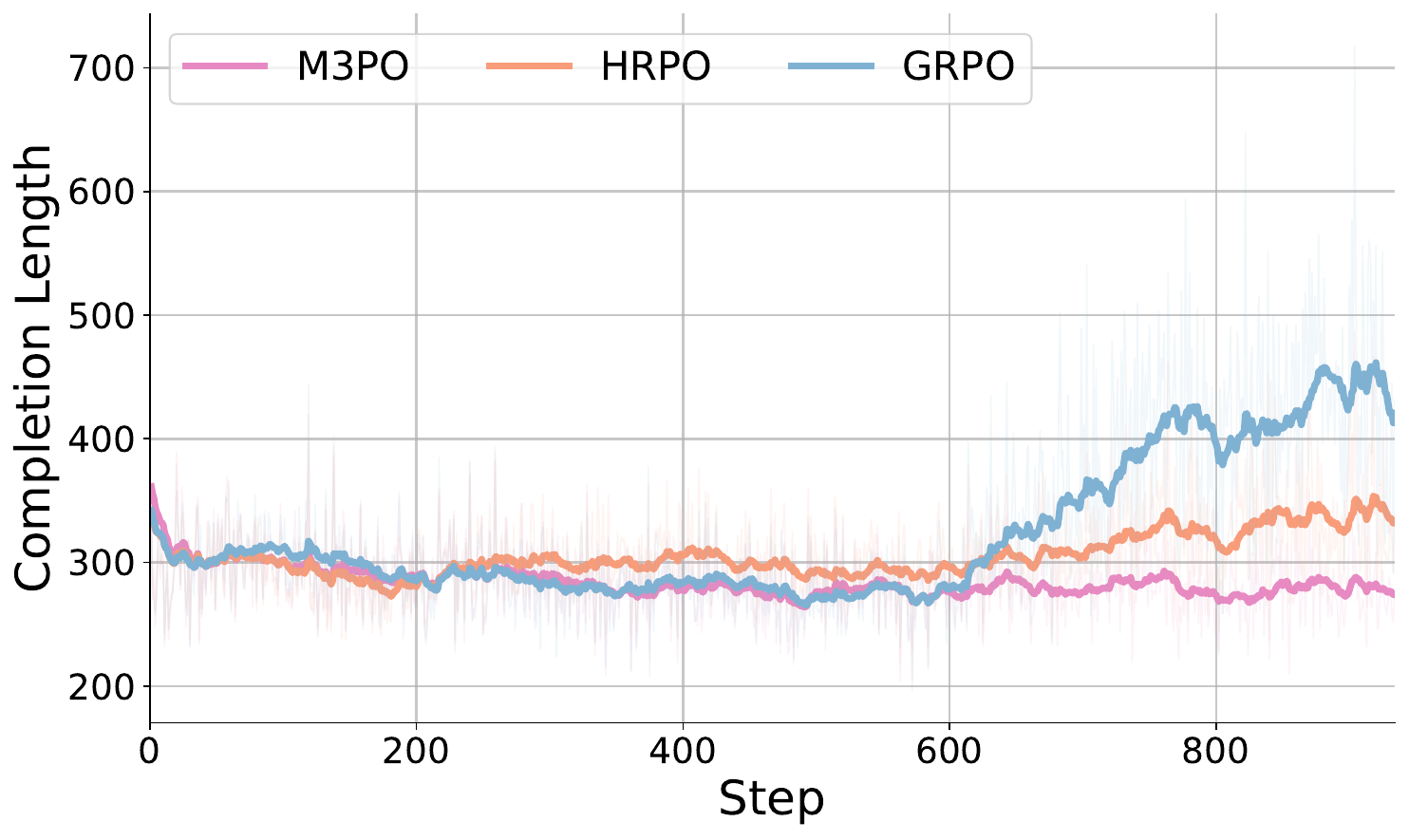}
    }
    \caption{Comparison of reward and completion length across training runs on GSM8k with the Qwen-1.5B backbone. M3PO exhibits the best training stability while producing the shortest completions.}
    \label{fig:1.5B_gsm8k}
\end{figure*}

\begin{figure*}[htbp]
    \centering
    \captionsetup[subfigure]{margin=2pt}
    % 设置子标题编号的位置是120pt
    \subfloat[Reward.]{
        \label{subfig:acc_plot}
        \centering
    \includegraphics[width=0.48\linewidth]{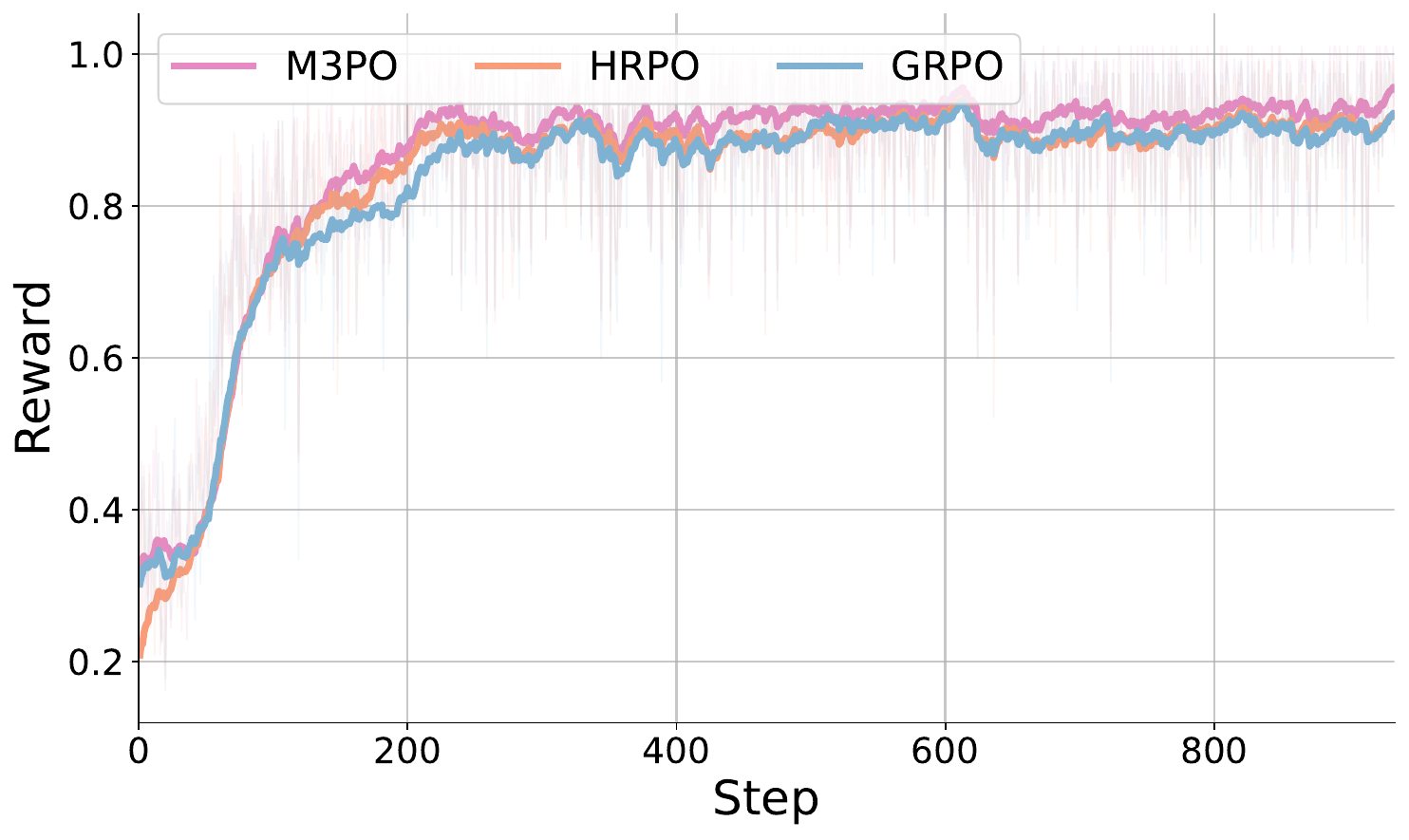}
    }
    \hspace{2pt}
    \subfloat[Completion length.]{
            \label{subfig:throughput_plot}
        \centering
    \includegraphics[width=0.48\linewidth]{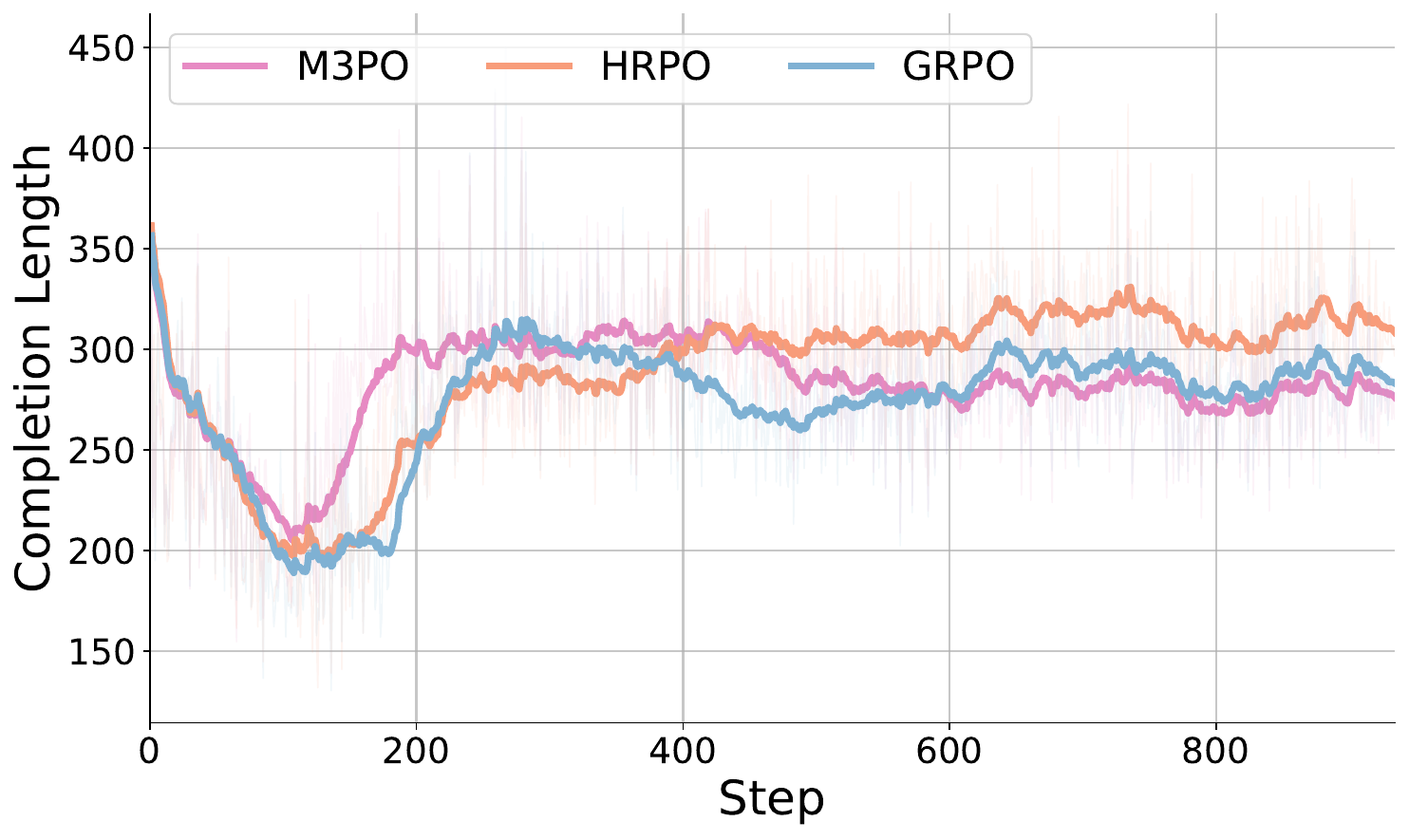}
    }
    \caption{Comparison of reward and completion length across training runs on GSM8k using the Qwen-3B backbone. M3PO achieves the highest reward while generating the shortest completions.}
    \label{fig:3B_gsm8k}
\end{figure*}

\begin{figure*}[htbp]
    \centering
    \captionsetup[subfigure]{margin=2pt}
    % 设置子标题编号的位置是120pt
    \subfloat[Reward.]{
        \label{subfig:acc_plot}
        \centering
    \includegraphics[width=0.48\linewidth]{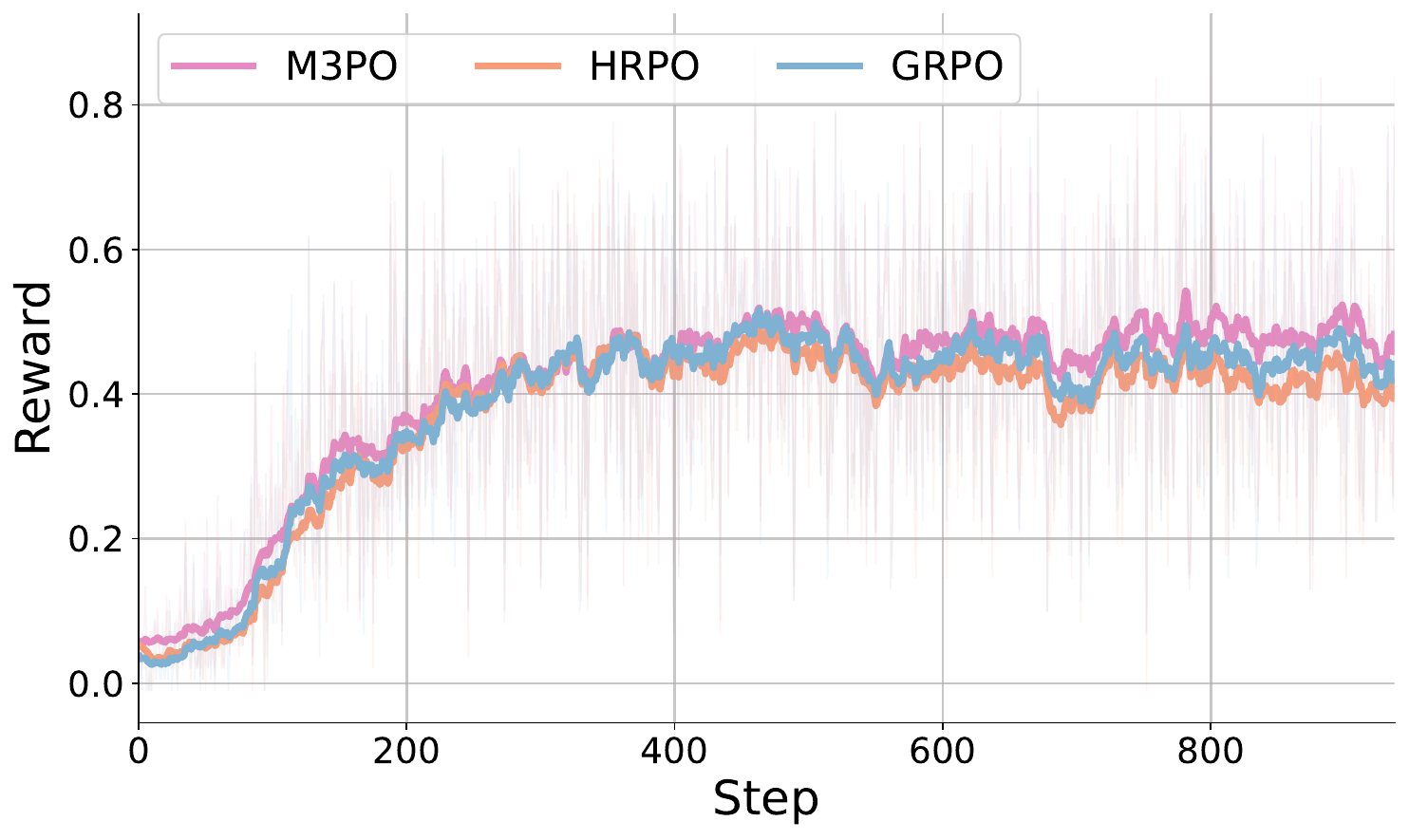}
    }
    \hspace{2pt}
    \subfloat[Completion length.]{
            \label{subfig:throughput_plot}
        \centering
    \includegraphics[width=0.48\linewidth]{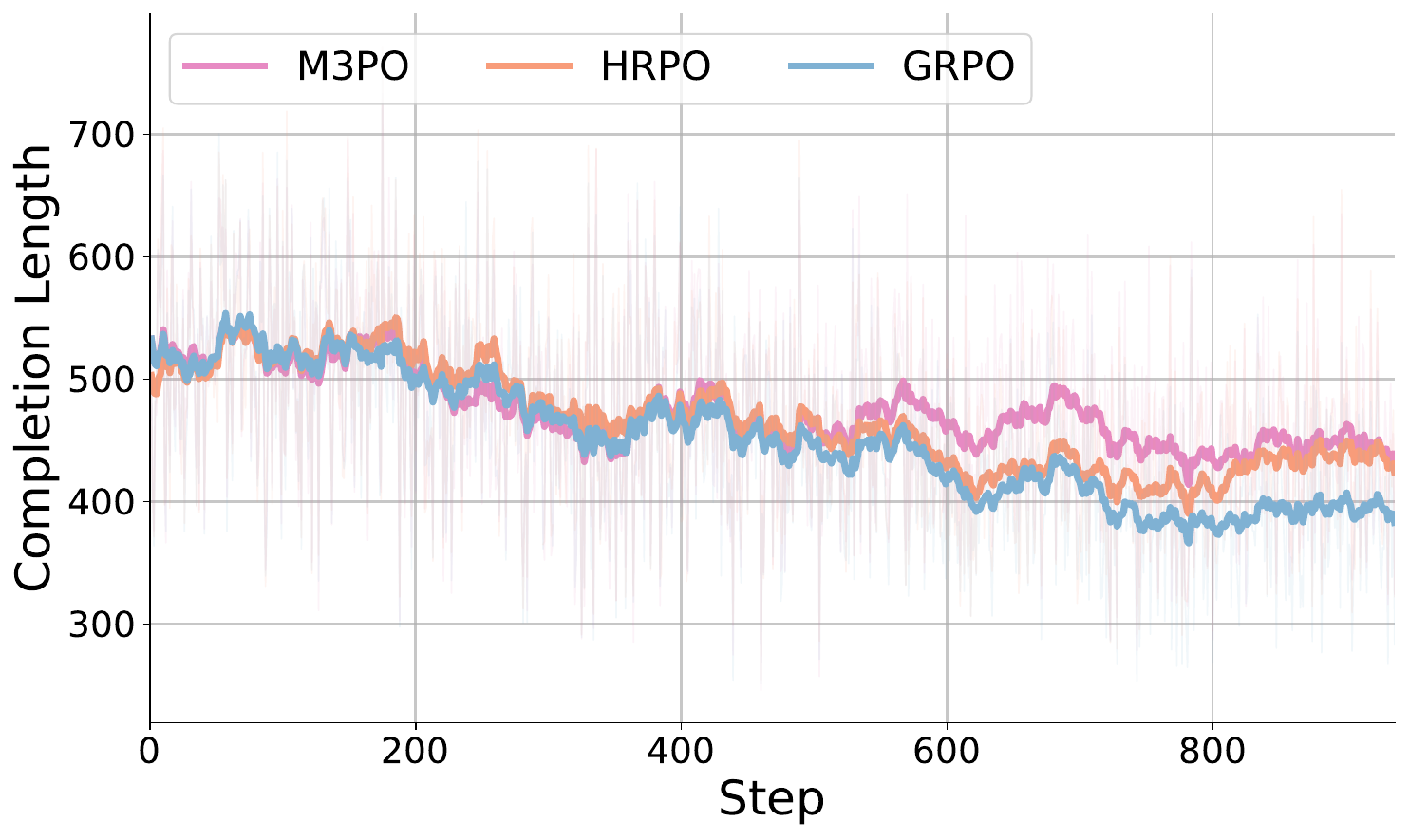}
    }
    \caption{Comparison of reward and completion length across training runs on MATH using the Qwen-1.5B backbone. M3PO attains the highest reward while maintaining comparable completion lengths.}
    \label{fig:1.5B_math}
\end{figure*}

\begin{figure*}[htbp]
    \centering
    \captionsetup[subfigure]{margin=2pt}
    % 设置子标题编号的位置是120pt
    \subfloat[Reward.]{
        \label{subfig:acc_plot}
        \centering
    \includegraphics[width=0.48\linewidth]{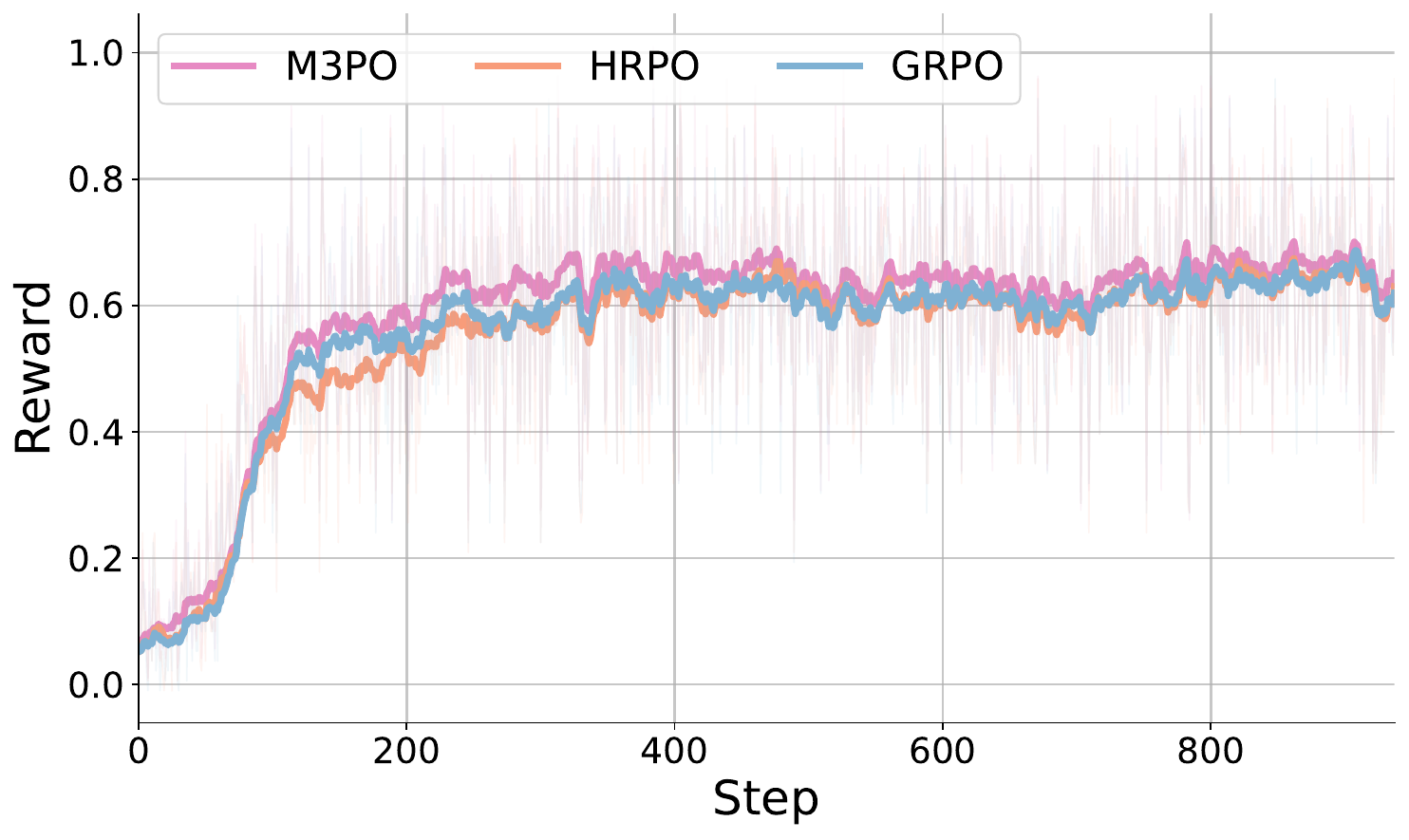}
    }
    \hspace{2pt}
    \subfloat[Completion length.]{
            \label{subfig:throughput_plot}
        \centering
    \includegraphics[width=0.48\linewidth]{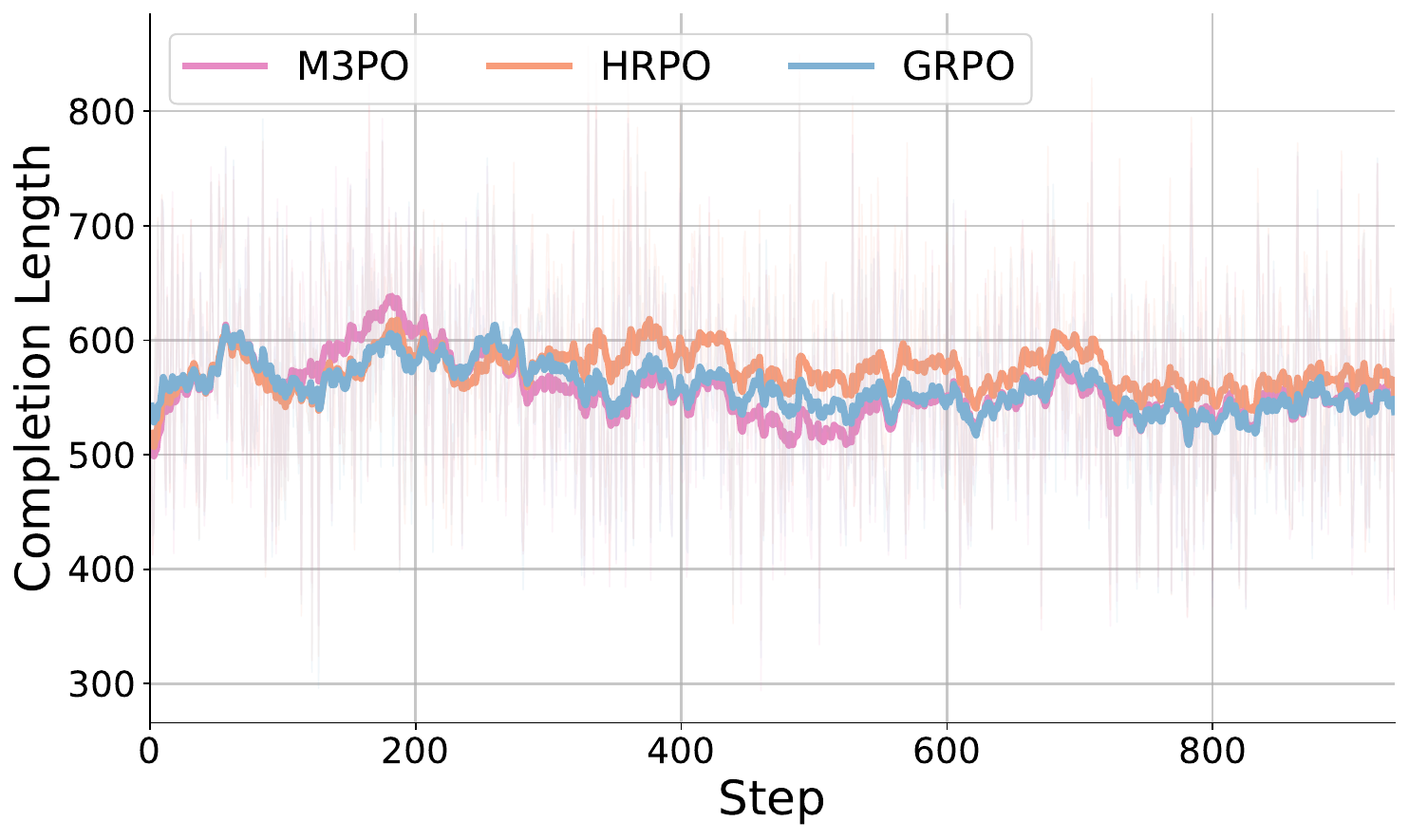}
    }
    \caption{Comparison of reward and completion length across training runs on MATH using the Qwen-3B backbone. M3PO achieves the highest reward while maintaining comparable completion length.}
    \label{fig:3B_math}
\end{figure*}

\paragraph{Efficiency analysis.} 
Figures~\ref{fig:1.5B_gsm8k}, \ref{fig:3B_gsm8k}, \ref{fig:1.5B_math}, \ref{fig:3B_math} present comprehensive comparisons of reward trajectories and reasoning chain lengths across GSM8k and MATH datasets using Qwen-1.5B and 3B model variants. On GSM8k, M3PO demonstrates exceptional training stability while achieving the highest final reward among all compared methods. This advantage is particularly pronounced with the Qwen-1.5B configuration, where M3PO maintains consistent convergence throughout the training process. In contrast, both HRPO and GRPO exhibit significant performance instability, characterized by a notable collapse around 600 training steps that substantially impacts their final performance. Beyond reward superiority, M3PO yields the most compact training completions, reducing unnecessary computational overhead while maintaining solution quality.

The MATH benchmark further validates M3PO's effectiveness, where it achieves superior performance with reasoning chain lengths comparable to other methods. This consistent pattern across datasets and model scales demonstrates M3PO's robust generalization capability and computational efficiency. These empirical results establish M3PO as an effective plug-and-play module that optimally balances performance with computational efficiency.

% To evaluate the efficiency of M3PO, we compare reasoning chain lengths across different latent reasoning approaches under identical experimental conditions using Qwen-3B backbone on MATH. As shown in Figure \ref{fig:latent_length}, both M3PO and HRPO exhibit an initial increase in completion length around step 200, followed by a progressive decrease until reaching a stable balance point. While all methods eventually converge to comparable lengths, M3PO achieves the most compact reasoning trajectories among all compared approaches.

Notably, the core architecture of M3PO remains parameter-efficient, introducing no additional structural parameters beyond the base model. This lightweight architecture contrasts with HRPO, which embeds trainable parameters directly into its reasoning architecture. Furthermore, M3PO preserves the same computational efficiency as standard autoregressive models during deployment, while delivering superior reasoning performance. These advantages establish M3PO as an effective and practical solution for robust reasoning.
% To examine the efficiency of M3PO, we first compare the completion length across various latent reasoning strategies under the same experimental settings, including Soft Thinking and HRPO. M3PO and HRPO shows a length increase at the 200 steps and gressivley decrease until find a balence point. It can be seen that these methods achieves the comparable completion length at the end, where our M3PO achievese the least length. During policy traiing, M3PO does not involve any additional parametwes and keep the ligheweight design. Moreover, compared to HRPO or Soft Thinking, which still need the soft aggreation during the inference, M3PO maintains the same effeicieny of the standard LLMs. It demonstartes that xxx

% Beyond the RL baselines examined in our main experiments, we further compare M3PO against representative latent reasoning approaches, including HRPO, COCONUT, and CODI. All methods are evaluated on the GSM8k and MATH benchmarks using the Qwen-1.5B backbone. For COCONUT, we utilize its augmented CoT training data, while for CODI we adopt the original CoT trajectories from each dataset. HRPO is implemented under the same settings as M3PO to ensure a fair comparison.
% \begin{figure}[htbp]
%     \centering
%     \includegraphics[width=1.0\linewidth]{figures/latent_reasoning_length.pdf}
%     \caption{Completion length comparison across different latent reasoning paradigms.}
%     \label{fig:latent_length}
% \end{figure}

% \section{Method Algorithm}

\section{Qualitative Analysis}
% 为了Qualitative证明我们方法的优越性，我们进行了Case study. 如图16所示，我们首先跟Soft Thinking进行了对比。可以看到， soft thinking存在很严重的格式，错误插值，逻辑不连贯等问题，而我们的M3PO中展现了非常干净，逻辑连贯，紧凑的推理过程。这证明了SOFT THINKG这种推理范式会引入噪音打破原先的推理连贯性，以及我们方法在xxx方法的优越性。

\begin{figure*}
    \centering
    \includegraphics[width=1.0\linewidth]{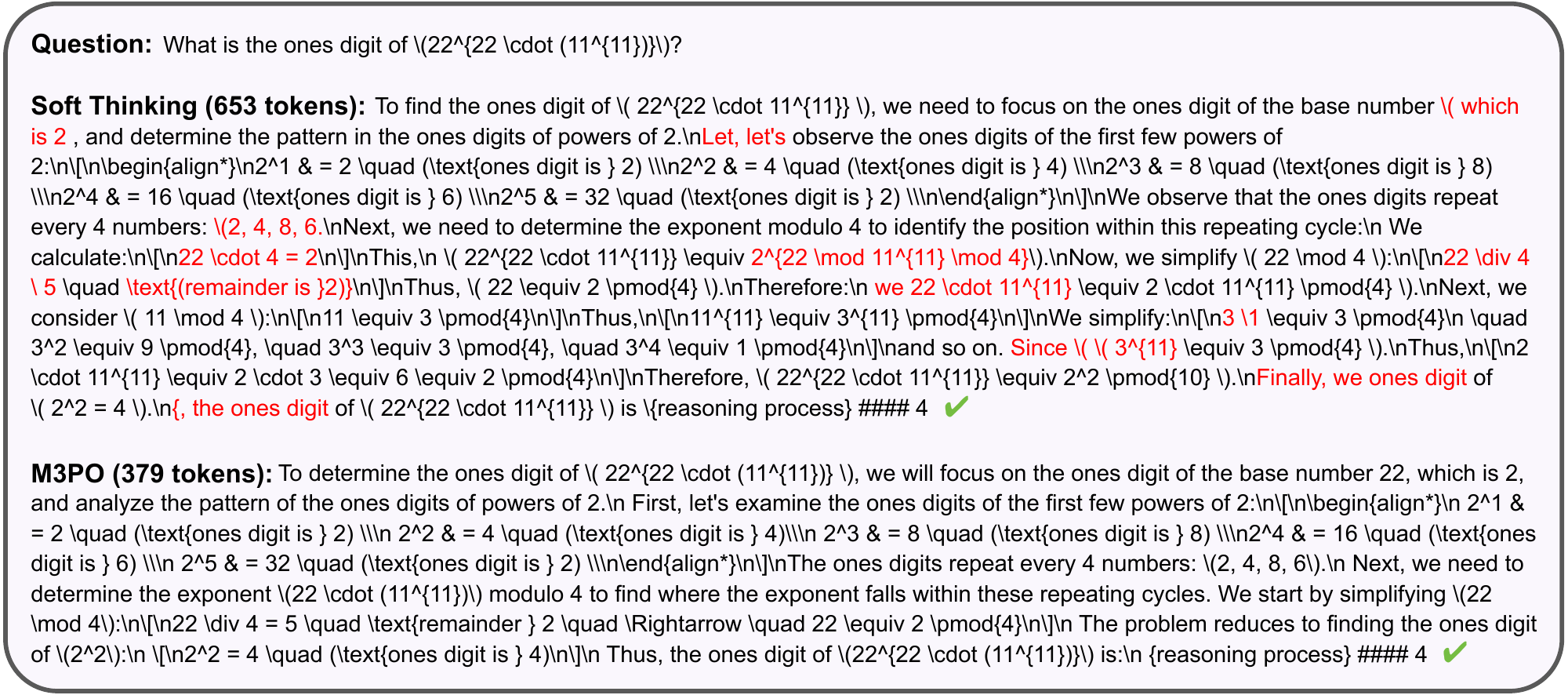}
    \caption{Case study comparing Soft Thinking and M3PO on a mathematical problem. Text highlighted in \textcolor{red}{red} indicates erroneous insertions, discontinuities, or formatting defects. Soft Thinking shows noisy, spurious text, whereas M3PO presents clean, consistently formatted steps.
. }
    \label{fig:case_m3po_st}
\end{figure*}

% 图17展示了M3PO和HRPO的case study. 可以看到HRPO表现出了复读机现象，epetitive loops that persist to the maximum completion length。而我们的方法展现出了整洁，逻辑连贯的推理形式。这证明了虽然我们和HRPO都属于混合推理范式，但是我们的多路径协同范式表现出更加鲁棒的推理能力。xxxx
To qualitatively demonstrate the advantages of our approach, we conduct detailed case studies comparing M3PO with alternative reasoning methods. Figure~\ref{fig:case_m3po_st} presents a comparison between M3PO and Soft Thinking. The Soft Thinking approach exhibits significant formatting issues, erroneous token insertions, and logical discontinuities throughout its reasoning chain. In contrast, M3PO generates a clean, logically coherent, and compact reasoning process free from these artifacts. This comparison clearly demonstrates how continuous soft aggregation in Soft Thinking introduces noise that disrupts reasoning coherence, while highlighting M3PO's superior ability to maintain structured reasoning patterns.

\begin{figure*}
    \centering
    \includegraphics[width=1.0\linewidth]{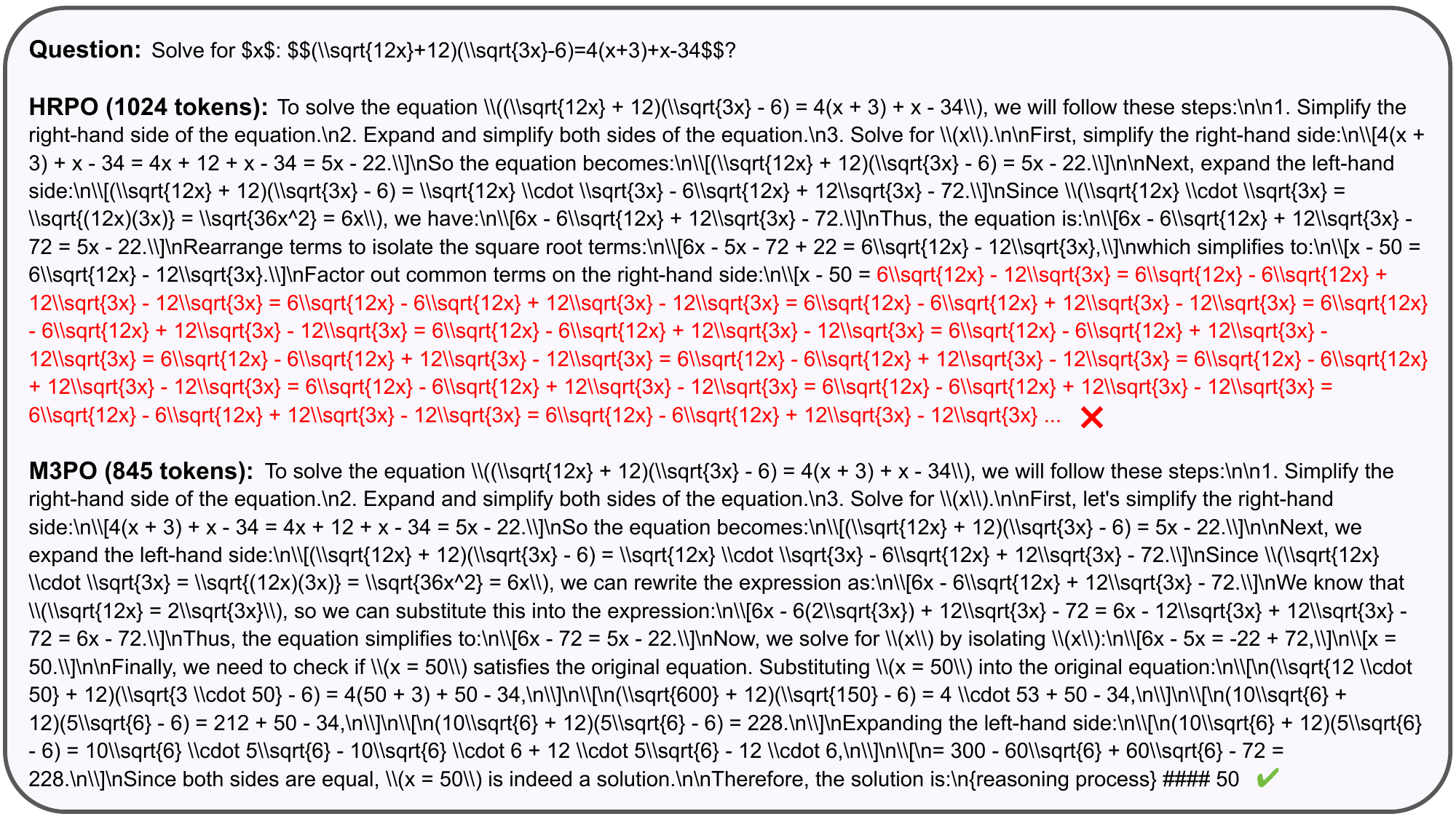}
    \caption{Case study comparing HRPO and M3PO on a mathematical problem. Text in \textcolor{red}{red} denotes erroneous insertions, discontinuities, or formatting defects. HRPO exhibits repetitive loops that persist to the maximum completion length, whereas M3PO produces coherent, logically structured steps. }
    \label{fig:case_m3po_hrpo}
\end{figure*}

Figure~\ref{fig:case_m3po_hrpo} provides a case study comparing M3PO with HRPO. While both methods employ hybrid reasoning paradigms, HRPO displays noticeable repetitive looping behavior that persists to the maximum completion length without reaching a solution. In contrast, M3PO correctly solves the problem through a well-structured and logically consistent reasoning trajectory. This qualitative difference demonstrates M3PO's superior robustness in maintaining coherent reasoning progress without degenerative patterns.

\begin{figure*}
    \centering
    \includegraphics[width=1.0\linewidth]{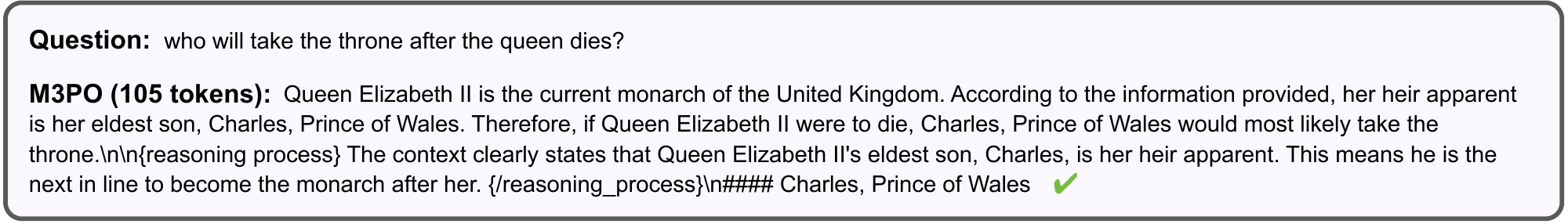}
    \caption{Reasoning example 1: M3PO on a knowledge-intensive task. The context is omitted for brevity.}
    \label{fig:example1}
\end{figure*}

\begin{figure*}
    \centering
    \includegraphics[width=1.0\linewidth]{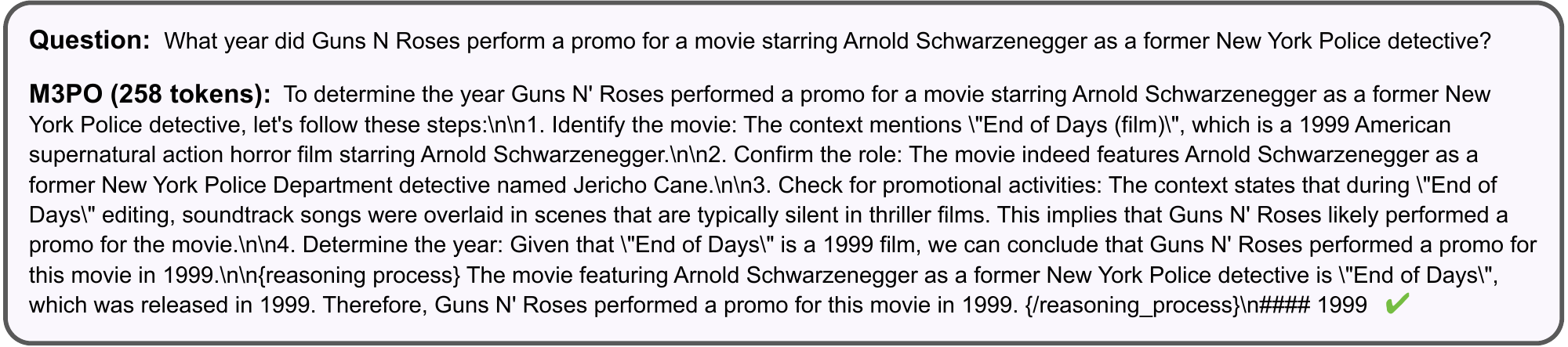}
    \caption{Reasoning example 2: M3PO on a knowledge-intensive task. The context is omitted for brevity.}
    \label{fig:example2}
\end{figure*}

\begin{figure*}
    \centering
    \includegraphics[width=1.0\linewidth]{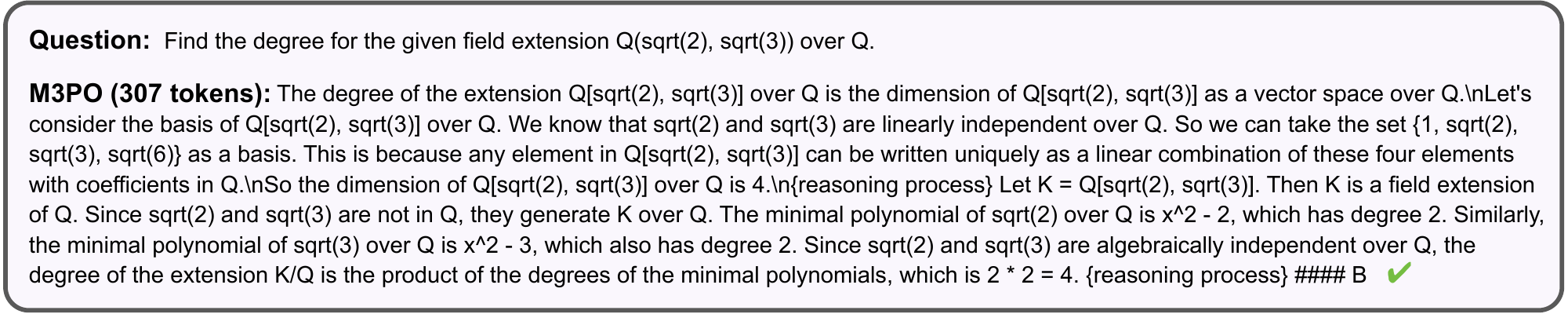}
    \caption{Reasoning example 3: M3PO on a MMLU-STEM task. The options are omitted for brevity.}
    \label{fig:example3}
\end{figure*}

\begin{figure*}
    \centering
    \includegraphics[width=1.0\linewidth]{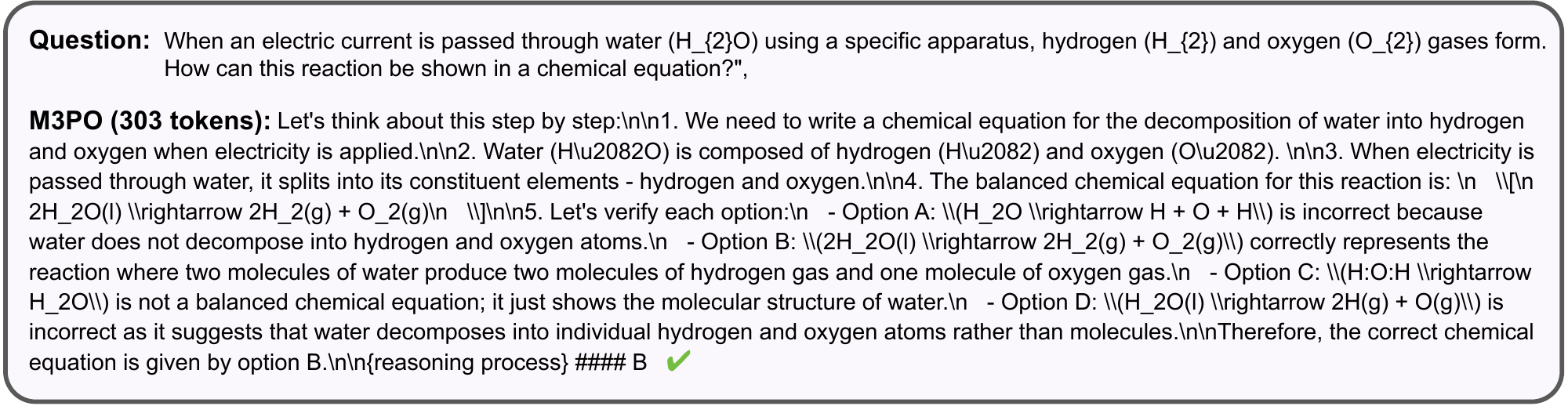}
    \caption{Reasoning example 4: M3PO on an ARC-Challenge task. The options are omitted for brevity.}
    \label{fig:example4}
\end{figure*}

\begin{figure*}
    \centering
    \includegraphics[width=1.0\linewidth]{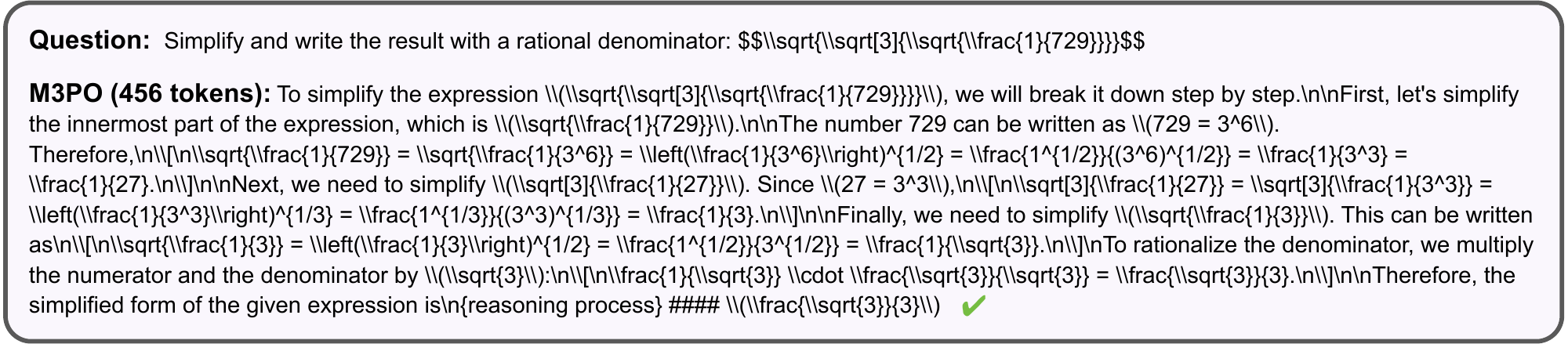}
    \caption{Reasoning example 5: M3PO on a mathematical task.}
    \label{fig:example5}
\end{figure*}

\begin{figure*}
    \centering
    \includegraphics[width=1.0\linewidth]{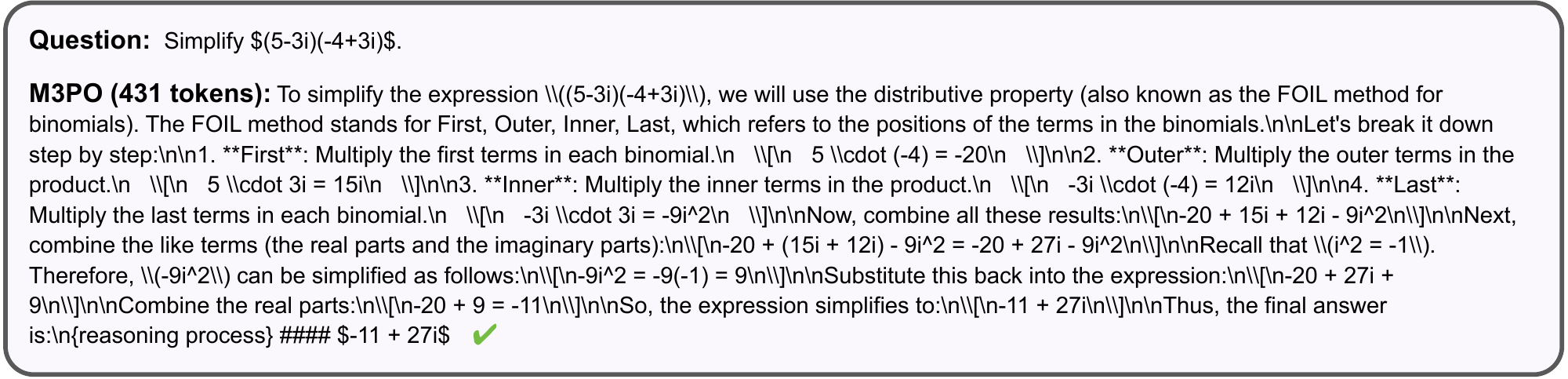}
    \caption{Reasoning example 6: M3PO on a mathematical task. }
    \label{fig:example6}
\end{figure*}

Figures \ref{fig:example1} to \ref{fig:example6} present reasoning examples generated by M3PO across diverse domains, including knowledge-intensive tasks, scientific reasoning, and mathematical problems. These cases collectively demonstrate M3PO's consistent ability to produce readable, coherent, and compact reasoning chains that lead to correct solutions.

In knowledge-intensive domains (Figures \ref{fig:example1} and \ref{fig:example2}), M3PO integrates factual information with logical inference, maintaining clarity throughout multi-step reasoning processes. For scientific reasoning tasks (Figures \ref{fig:example3} and \ref{fig:example4}), the method demonstrates precise conceptual understanding and systematic problem-solving approaches. Mathematical problems (Figures \ref{fig:example5} and \ref{fig:example6}) further showcase M3PO's capacity for structured derivation and computational accuracy.

Across all domains, M3PO generates reasoning chains characterized by logical coherence and expressive clarity. This behavior is a direct consequence of our multi-path coordination mechanism that implicitly guides gradient updates during policy optimization, enabling the model to develop robust reasoning patterns without compromising output coherence. This qualitative evidence complements our quantitative results, demonstrating the practical utility of our approach for generating interpretable and reliable reasoning across different task types.

\end{document}